# *SOLD*: Sinhala Offensive Language Dataset

Tharindu Ranasinghe[1*], Isuri Anuradha[2], Damith Premasiri[2], Kanishka Silva[2], Hansi Hettiarachchi[3], Lasitha Uyangodage[4] and Marcos Zampieri[5]

[1]Aston University, Birmingham, UK.
[2]University of Wolverhampton, Wolverhampton, UK.
[3]Birmingham City University, Birmingham, UK.
[4]University of Münster, Münster, Germany.
[5]George Mason University, Fairfax, VA , USA.

*Corresponding author(s). E-mail(s): t.ranasinghe@aston.ac.uk;

**Abstract**

The widespread of offensive content online, such as hate speech and cyber-bullying, is a global phenomenon. This has sparked interest in the artificial intelligence (AI) and natural language processing (NLP) communities, motivating the development of various systems trained to detect potentially harmful content automatically. These systems require annotated datasets to train the machine learning (ML) models. However, with a few notable exceptions, most datasets on this topic have dealt with English and a few other high-resource languages. As a result, the research in offensive language identification has been limited to these languages. This paper addresses this gap by tackling offensive language identification in Sinhala, a low-resource Indo-Aryan language spoken by over 17 million people in Sri Lanka. We introduce the Sinhala Offensive Language Dataset (*SOLD*) and present multiple experiments on this dataset. *SOLD* is a manually annotated dataset containing 10,000 posts from Twitter annotated as offensive and not offensive at both sentence-level and token-level, improving the explainability of the ML models. *SOLD* is the first large publicly available offensive language dataset compiled for Sinhala. We also introduce *SemiSOLD*, a larger dataset containing more than 145,000 Sinhala tweets, annotated following a semi-supervised approach.

⚠This paper contains texts that may be offensive and harmful.

# 1 Introduction

Offensive posts on social media platforms result in a number of undesired consequences to users. They have been investigated as triggers of suicide attempts, and ideation [1, 2], and mental health conditions such as depression [3, 4]. Content moderation in online platforms is often applied to mitigate these serious repercussions. As human moderators cannot cope with the volume of posts online, there is a need for automatic systems that can assist them [5]. Social media platforms have been investing heavily in developing these systems and several studies in NLP have been conducted to tackle this problem [6]. Most studies proposed a supervised approach to detect offensive content automatically using various models ranging from traditional ML approaches to more recent neural-based methods trained on language-specific annotated data [7].

Considering the importance of annotated data, there is a growing interest in the NLP community to develop datasets that are capable of training ML models to detect offensive language. These datasets focus on various kinds of offensive content such as abuse [5, 8, 9], aggression [10, 11], cyber-bullying [12, 13], toxicity [14, 15], and hate speech [16, 17]. Furthermore, competitive shared tasks such as OffensEval [18, 19], TRAC [20, 21], HASOC [22, 23], HatEval [24] and AbuseEval [25] have created various benchmark datasets on the topic. Apart from a few notable exceptions, the majority of these datasets are built on English [26] and other high-resource languages such as Arabic [27, 28], Danish [29], Dutch [30] and French [31]. However, offensive language in social media is not limited to specific languages. Most popular social media platforms, such as Twitter and Facebook, are highly multilingual, as users express themselves in their mother tongue [32]. There is a considerable urgency to address offensive speech in different languages, but the lack of annotated datasets limits offensive language identification in low-resource languages.

In this paper, we revisit the task of offensive language identification for low-resource languages. Our work focuses on Sinhala, an Indo-Aryan language spoken by over 17 million people in Sri Lanka. Sinhala is one of the two official languages in Sri Lanka. Most of the people who speak Sinhala are the Sinhalese people of Sri Lanka, who make up the largest ethnic group on the island. Even though Sinhala is spoken by a large population, it is relatively low-resourced compared to other languages spoken in the region. According to Kepios analysis[1] the number of social media users in Sri Lanka at the start of 2022 was equivalent to 38.1% of the total population with users increasing by 300,000 between 2021 and 2022. Despite this growth, the spread of offensive posts on social media platforms is still a huge and largely unaddressed concern in Sri Lanka. In 2019, after the Easter bombings that targeted Christian churches[2], the government had to temporarily block all the social media on the island to curtail the spread of hate speech against Muslims. Similarly, both in 2019 and 2022, the government again blocked all social media platforms on the island to control offensive speech against the government[3]. These widespread social media bans not only violate rights to free speech but also limit the general public's accessibility to authorities and health services in dire situations. Therefore, a system that can detect offensive posts and help content moderators to remove them is paramount in Sri Lanka, and we believe the datasets and research presented in this paper will be the first step toward this goal.

---

[1] A complete analysis is available on https://datareportal.com/reports/digital-2022-sri-lanka
[2] https://www.bbc.com/news/world-asia-48010697

We collect and annotate data from Twitter to create the largest Sinhala offensive language identification dataset to date; *SOLD*. We first annotate the tweets at the sentence level for offensive/ not-offensive labels. Most offensive language identification datasets follow a similar approach and classify whole posts. However, identifying the specific tokens that make a text offensive can assist human moderators and contribute to building more explainable models for offensive language identification. Explainable ML is a widely discussed topic in the NLP community [33]. Several English offensive language datasets have been annotated at token-level [34, 35] to support explainability. Following this, we annotate *SOLD* both at the post and at the token-level. If a text is offensive, we label each token based on its contribution to the overall offensiveness at the sentence level. If the token adds to the offensiveness, it is annotated as offensive; otherwise, it is marked as not offensive. As far as we know, *SOLD* is the first non-English offensive language detection dataset with explainable tokens.

Data scarcity is a major challenge in building ML models for low-resource languages like Sinhala [36]. In this paper, we explore two approaches to overcome data scarcity. 1. We perform **transfer learning**. We draw inspiration from recent work that applied cross-lingual models for low-resource offensive language identification [7] and adapted it to Sinhala. 2. We perform **semi-supervised data augmentation**. Motivated by SOLID [37], the largest offensive language dataset available for English, we propose a similar semi-supervised data augmentation approach for Sinhala. We collect more than 145,000 Sinhala tweets and annotate them using a semi-supervised approach. We release the resource as *SemiSOLD* and use it to improve Sinhala offensive language detection results. As far as we know, *SemiSOLD* is the largest non-English offensive language online dataset annotated in a semi-supervised manner. We believe that the findings of these two approaches will benefit many low-resource languages.

We summarise our contributions in this paper as follows,

1. We release *SOLD*[4], the largest annotated Sinhala Offensive Language Dataset to date. *SOLD* contains 10,000 annotated tweets for offensive language identification at sentence-level and token-level.
2. We experiment with several machine learning models, including state-of-the-art transformer models, to identify the offensive language at sentence-level and token-level. To the best of our knowledge, the identification of offensive language at both sentence-level and token-level has not been attempted on Sinhala.
3. We explore offensive language identification with cross-lingual embeddings and transfer learning. We take advantage of existing data in high-resource languages such as English to project predictions to Sinhala. We show that transfer learning can improve the results in Sinhala, which could benefit many low-resource languages.
4. We investigate semi-supervised data augmentation. We create *SemiSOLD*; a larger semi-supervised dataset with more than 145,000 instances for Sinhala. We use multiple machine learning models trained on the annotated training set and combine the scores following a similar methodology described in [37]. We show that this semi-supervised dataset can be used to augment the training set, which improves the results of machine learning models.
5. Finally, we demonstrate the explainability of the sentence-level offensive language identification models in Sinhala using token-level annotations in *SOLD*. We experiment with how transfer learning and semi-supervised data augmentation affect the explainability of the models. To the best of our knowledge, the explainability of the offensive language models has not been explored in low-resource languages.

---

[3]https://www.bbc.co.uk/news/technology-48022530
[4]Dataset available at: https://github.com/Sinhala-NLP/SOLD

With these resources released in this paper, we aim to answer the following research questions:

- **RQ1– Performance**: How do the state-of-the-art machine learning models perform in Sinhala offensive language identification at sentence-level and token-level?
- **RQ2– Data scarcity**: Our second research question addresses data scarcity, a known challenge for low-resource NLP. We divide it into two parts as follows:
  - **RQ2.1**: Do available resources from resource-rich languages combine with transfer-learning techniques aid the detection of offensive language in Sinhala at sentence-level and token-level?
  - **RQ2.2**: Can semi-supervised data augmentation improve the results for Sinhala offensive language identification at sentence-level?
- **RQ3– Explainability**: Our third research question addresses the explainability of the machine learning models, a topic of interest for the NLP community, yet not explored in low-resource languages. We divide it into three parts as follows:
  - **RQ3.1**: How to demonstrate explainability of the sentence-level offensive language identification using token-level annotations in Sinhala?
  - **RQ3.2**: Does transfer-learning from resource-rich languages affect the explainability of the offensive language identification models?
  - **RQ3.3**: Can semi-supervised data augmentation improve the explainability?

Finally, the development of *SOLD* and *SemiSOLD* open exciting new avenues for research in Sinhala offensive language identification. We train a number of state-of-the-art computational models on this dataset and evaluate the results in detail, making this paper the first comprehensive evaluation of Sinhala offensive language online. The rest of the paper is organised as follows. Section 2 highlights the recent research in offensive language identification. Section 3 describes the data collection, annotation process and statistical analysis of the dataset. Section 5 presents the experiments at both sentence-level and token-level. Sections 6 and 7 show the various transfer learning and semi-supervised data augmentation techniques we employed, respectively. Section 8 summarises the conclusions of this study revisiting the above RQs.

## 2 Related Work

The problem of offensive content online continues to attract attention within the AI and NLP communities. In recent studies, researchers have created datasets and trained various systems to identify offensive content in social media. Popular international competitions on the topic have been organised at conferences such as OffensEval [18, 19], TRAC [20, 21], HASOC [22, 23], HatEval [24], and AbuseEval [25]. These competitions attracted many participants, and they provided participants with various important benchmark datasets, allowing them to train competitive systems on them [38, 39].

In terms of languages, due to the availability of annotated datasets, the vast majority of studies in offensive language identification use English [40, 41] and other high-resource languages such as Arabic [27, 28], Dutch [30], French [31], German [42], Greek [43], Italian [44], Portuguese [45], Korean [46], Slovene [47], Spanish [48] and Turkish [49]. More recently, several offensive language online datasets have been annotated on low-resource languages such as Bengali [50], Marathi [36, 51] Nepali [52], Tamazight [53], and Urdu [54]. These datasets have been annotated on coarse-grained labels such as offensive/ not offensive and hate speech/ non-hate speech. Some of these datasets have even been annotated on fine-grained labels. For example, offensive tweets in Urdu [54] have been further annotated as abusive, sexist, religious hate and profane, while the offensive tweets in Marathi [36] have been further annotated into targeted and untargeted offence. For Sinhala, too, there is a hate speech detection dataset [55] where Facebook posts have been annotated for three labels; hate, offensive and neutral speech detection. However, the dataset is limited in size as it contains only 3,000 posts, and the dataset is not publicly available. Another related Sinhala dataset for offensive language identification is Sinhala-CMCS [56], where 10,000 social media comments have been annotated for five classification tasks; sentiment analysis, humour detection, hate speech detection, language identification, and aspect identification. However, the dataset is based on Sinhala–English code-mixed texts. With the development of keyboards that support Sinhala script, such as Helakuru[5], there is an increasing number of social media users who use Sinhala script in their conversations. Therefore, a Sinhala offensive language identification dataset with Sinhala script is a research gap we address in this paper.

---
[5]Helakuru is a popular Android keyboard that supports typing in Sinhala script

All the datasets we mentioned before are sentence-level offensive language identification datasets where the whole sentence is given a single label. While sentence-level offensive language datasets have been popular in the community, identifying the specific tokens that make a text offensive can be useful in many applications [57]. Furthermore, token-level annotations can be used to improve the explainability of the sentence-level models [58, 59]. As a result, detecting tokens instead of entire posts has been studied in many domains, including propaganda detection [60] and translation error detection [61]. In the offensive language domain too, two datasets have been released with explainable token-level labels; HateXplain [34], and TSD [35]. Both of these datasets have sentence-level labels together with token-level labels. TSD dataset was released for the SemEval 2021 - Task 5 [62][6]. While token-level offensive language identification is an important research area, as far as we know, no non-English datasets have been annotated at the token-level. With *SOLD*, we hope to address this gap with token-level annotations, contributing to the first explainable non-English offensive language identification dataset.

In machine learning approaches, sentence-level offensive language identification has often been considered a text classification task [63, 64]. Early approaches utilised classical machine learning classifiers such as SVMs with feature engineering [65] to perform sentence-level offensive language identification. With the introduction of word embeddings, [66, 67], different neural network architectures were used to perform offensive language identification [68]. These architectures contain different techniques such as long short-term memory networks [69, 70], convolutional neural networks [71, 72], capsule networks [73, 74] and graph convolutional networks [75]. With the recent development of large pre-trained transformer models such as BERT [76], and XLNET [77], several studies have explored the use of general pre-trained transformers by fine-tuning them in sentence-level offensive language tasks [39, 78]. These approaches have provided excellent results and outperformed previous architectures in many datasets [38, 79, 80]. Going beyond fine-tuning, recent approaches such as fBERT [81] and HateBERT [82] have trained domain-specific transformer models on offensive language corpora which have provided state-of-the-art results in many benchmarks. Finally, multilingual pre-trained transformer models such as mBERT [76] and XLM-RoBERTa [83] have enabled cross-lingual transfer learning, which makes it possible to leverage available English resources to make predictions in languages with fewer resources helping to cope with data scarcity in low-resource languages [7, 32, 84].

Token-level offensive language identification has been commonly addressed as a token classification task where a machine learning model will predict whether each token is offensive or not [85]. Besides machine learning models, researchers have explored lexicon-based approaches too [86, 87]. Three kinds of lexicon-based methods have been used in the past; 1. Lexicon was handcrafted by domain experts and was simply employed as a list of toxic words for lookup operations [87]. 2. Lexicon was compiled using the set of tokens labelled as positive (offensive, toxic etc.) in the training set, and it was used as a lookup table [88]. 3. Supervised lexicons were built with statistical analysis on the occurrences of tokens in a training set solely annotated at the sentence-level [89]. While lexicon-based approaches provide simple solutions, they are usually outperformed by machine learning approaches [86]. Therefore, they have been merely used as baselines. Many deep learning architectures have been explored at the token-level too. Long short-term memory networks [57, 90] and convolutional neural networks [91, 92] have been popular among them. Similar to sentence-level offensive language detection, pre-trained transformer models such as BERT [76] and XLNET [77] have provided state-of-the-art results in token-level. These approaches either use the default token classification architecture in transformers [93, 94] or use a conditional random field layer [95] on top of the transformer model [96, 97]. Based on this supervised learning paradigm, several open-source frameworks such as MUDES [93] have been released to perform token-level offensive language identification.

---

[6]More details about the toxic spans detection task can be found at https://sites.google.com/view/toxicspans

In addition to supervised approaches, researchers have explored weakly supervised approaches in token-level offensive language identification [34, 35] as it can be seen as a case of rationale extraction [98, 99]. These approaches use an attentional binary classifier to predict the sentence label and then invoke its attention at inference time to obtain offensive tokens as in rationale extraction. This allows leveraging existing training datasets that provide gold labels indicating sentence-level without providing gold labels at token-level. In recent years, researchers have explored various techniques such as attention scores of a long short-term memory classifier [90], long short-term memory classifier with a token-masking approach [89], SHAP [100] with a sentence-level fine-tuned transformer model [90] and combine LIME [58] with a sentence-level classifier [101]. All the approaches mentioned above used a threshold to turn the tokens' explanation scores (e.g., attention or LIME scores) into binary decisions (offensive/not-offensive tokens) [34, 102]. Although token classification approaches performed overall better, these approaches have performed surprisingly well, too, despite having been trained on data without token-level annotations [34, 35]. They have further contributed to explainable machine learning in offensive language identification.

All the token-level methods mentioned above have been experimented only with English data. With *SOLD*, we fill this gap by evaluating how these token-level offensive language detection methods perform in a low-resource language setting. Furthermore, due to the lack of suitable datasets, techniques we observed at the sentence-level, such as cross-lingual transfer learning and data augmentation, have not been explored widely at the token-level. In this paper, we will analyse the effect of transfer learning and data augmentation at the token-level for the first time.

# 3 Data Collection and Annotation

In the following subsections, we describe the data collection and annotation process of *SOLD*.

## 3.1 Data Collection

We retrieved the instances in *SOLD* from Twitter using its API[7] and Tweepy Python library[8]. We collect data by using predefined keywords, which is a common method in offensive language detection dataset construction [103, 104]. As keywords, we use words that are often included in offensive tweets such as *"you"* (තෝ, උඹ) and *"go"* (පලයන්, පල). We also include anti-government (@NewsfirstSL) and pro-government (@adaderanasin) news accounts. The complete list of keywords that were used to collect *SOLD* is shown in Table 2. However, Sinhala is written in three ways in social media. (a) Sinhala written in Sinhala script (b) Sinhala written in Roman script, pronunciation-based and (c) Mixed script text that contains both Sinhala and Roman scripts. Since our goal is to construct a Sinhala offensive language identification dataset in Sinhala script, we use TwitterAPI's language filter to have the tweets only written with Sinhala script. Using these keywords and the filtering strategy, we collected 10,500 tweets.

We do not collect Twitter user IDs to remove the users' personally identifiable information. We replace mentions of the usernames in the tweet with @USER tokens and URLs with <URL> tokens to conceal private information using regular expressions.

## 3.2 Annotation Task Design

We use an annotation scheme split into two levels deciding (a) Offensiveness of a tweet (sentence-level) and (b) Tokens that contribute to the offence at sentence-level (token-level). as shown in Figure 1. In the following section, we provide the definitions of sentence-level and token-level offensive language identification and the guidelines for each annotation task.

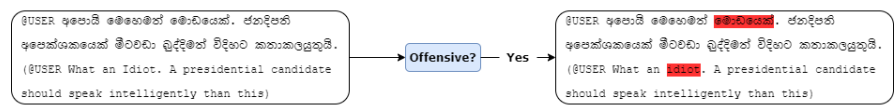

**Figure 1**: A translated example of SOLD. If an annotator marked a tweet as offensive, he/she was asked to highlight which tokens of the tweet justifies their decision.

---

[7]Twitter API is available at https://developer.twitter.com/en/docs/twitter-api/tools-and-libraries/v2

[8]Tweepy is an easy-to-use Python library for accessing the Twitter API available at https://www.tweepy.org/

|   | 0 | 1 | 2 | 3 | 4 | 5 | 6 | 7 | 8 | 9 |
|---|---|---|---|---|---|---|---|---|---|---|
| 0 | 1.00 | 0.74 | 0.82 | 0.85 | 0.77 | 0.73 | 0.80 | 0.77 | 0.87 | 0.78 |
| 1 | 0.74 | 1.00 | 0.77 | 0.74 | 0.67 | 0.81 | 0.86 | 0.77 | 0.74 | 0.77 |
| 2 | 0.82 | 0.76 | 1.00 | 0.84 | 0.78 | 0.82 | 0.75 | 0.77 | 0.68 | 0.71 |
| 3 | 0.86 | 0.74 | 0.84 | 1.00 | 0.78 | 0.80 | 0.75 | 0.79 | 0.73 | 0.72 |
| 4 | 0.77 | 0.66 | 0.78 | 0.78 | 1.00 | 0.77 | 0.66 | 0.76 | 0.76 | 0.83 |
| 5 | 0.73 | 0.81 | 0.82 | 0.80 | 0.77 | 1.00 | 0.83 | 0.66 | 0.64 | 0.71 |
| 6 | 0.80 | 0.86 | 0.75 | 0.75 | 0.66 | 0.83 | 1.00 | 0.78 | 0.74 | 0.73 |
| 7 | 0.77 | 0.77 | 0.77 | 0.79 | 0.75 | 0.66 | 0.78 | 1.00 | 0.75 | 0.74 |
| 8 | 0.88 | 0.78 | 0.68 | 0.72 | 0.76 | 0.64 | 0.73 | 0.74 | 1.00 | 0.79 |
| 9 | 0.78 | 0.75 | 0.76 | 0.74 | 0.83 | 0.76 | 0.74 | 0.75 | 0.79 | 1.00 |

(a) Inter Annotator Agreement in *SOLD* sentence-level

|   | 0 | 1 | 2 | 3 | 4 | 5 | 6 | 7 | 8 | 9 |
|---|---|---|---|---|---|---|---|---|---|---|
| 0 | 1.00 | 0.53 | 0.58 | 0.61 | 0.53 | 0.58 | 0.57 | 0.59 | 0.53 | 0.60 |
| 1 | 0.69 | 1.00 | 0.63 | 0.73 | 0.74 | 0.64 | 0.65 | 0.61 | 0.62 | 0.61 |
| 2 | 0.83 | 0.60 | 1.00 | 0.90 | 0.82 | 0.68 | 0.57 | 0.59 | 0.72 | 0.89 |
| 3 | 0.64 | 0.59 | 0.60 | 1.00 | 0.66 | 0.56 | 0.57 | 0.61 | 0.63 | 0.65 |
| 4 | 0.60 | 0.60 | 0.52 | 0.67 | 1.00 | 0.61 | 0.64 | 0.67 | 0.65 | 0.58 |
| 5 | 0.61 | 0.61 | 0.71 | 0.84 | 0.84 | 1.00 | 0.60 | 0.66 | 0.80 | 0.62 |
| 6 | 0.93 | 0.60 | 0.77 | 0.81 | 0.72 | 0.69 | 1.00 | 0.63 | 0.65 | 0.67 |
| 7 | 0.75 | 0.62 | 0.62 | 0.80 | 0.70 | 0.54 | 0.57 | 1.00 | 0.68 | 0.63 |
| 8 | 0.56 | 0.62 | 0.56 | 0.70 | 0.55 | 0.54 | 0.61 | 0.68 | 1.00 | 0.52 |
| 9 | 0.56 | 0.67 | 0.61 | 0.63 | 0.81 | 0.63 | 0.69 | 0.60 | 0.73 | 1.00 |

(b) Inter Annotator Agreement in *SOLD* token-level

**Figure 2**: Inter Annotator Agreement in *SOLD*.

### *Sentence-level Offensive Language*

Our sentence-level offensive language detection follows level A in OLID[104]. We asked annotators to discriminate between the following types of tweets:

- **Offensive (OFF)**: Posts containing any form of non-acceptable language (profanity) or a targeted offence, which can be veiled or direct. This includes insults, threats, and posts containing profane language or swear words.
- **Not Offensive (NOT)**: Posts that do not contain offense or profanity.

Each tweet was annotated with one of the above labels, which we used as the labels in sentence-level offensive language identification. Having broad offensive and not-offensive labels provides us with the opportunity to perform transfer learning as the majority of the offensive language datasets such as OLID [104] (English), OGDT (Greek) [43] and MOLD (Marathi) [51].

### *Token-level Offensive Language*

To provide a human explanation of labelling, we collect rationales for the offensive language. Following HateXplain [34], we define a rationale as a specific text segment that justifies the human annotator's decision of the sentence-level labels. Therefore, We ask the annotators to highlight particular tokens in a tweet that supports their judgement about the sentence-level label (offensive, not offensive). Specifically, if a tweet is offensive, we guide the annotators to highlight tokens from the text that supports the judgement while including non-verbal expressions such as emojis and morphemes that are used to convey the intention as well. These tokens can be used to train explainable models, as is shown in recent works [34, 35, 59].

### 3.3 Data Annotation

We follow prior work in the offensive language domain [26, 104, 105], and we annotate our data using crowd-sourcing. We used LightTag [106][9], a text annotation platform, to annotate the tweets. As hate speech annotation can be influenced by the bias of the annotators [26], we collected judgement from diverse annotators as possible. For the annotation task, we recruited a team of ten annotators. All of them are native Sinhala speakers, aged between 25-40, and everyone had at least a bachelor's degree qualification.

First, we provided the annotators with several in-person and virtual training sessions on LightTag. Once they completed them successfully, we first conducted a pilot annotation study followed by the main annotation task. In the pilot task, each annotator was provided with randomly selected 500 tweets from the collected dataset which had a similar keyword distribution. The annotators were required to do sentence-level annotations and token-level annotations if a tweet was annotated as offensive. To clearly understand the task, they were provided with multiple examples along with the annotation guidelines. The primary purpose of the pilot task was to collect feedback from the annotators to improve the annotation guidelines and the main annotation task. Furthermore, these annotations were used to ensure the balance between offensive and not-offensive classes.

After the pilot annotation, once we had improved the annotation guidelines, we started with the main annotation task. Since the pilot annotation showed that the offensive percentage of the dataset falls within our requirements, we did not collect further tweets or keywords. The main annotation task consisted of 10,000 tweets, that were not part of the pilot task. Each tweet was annotated by three annotators. To reduce the bias, we limit the maximum amount of annotation per person to 10% of the total annotations. Figure 2a shows the pairwise Fleiss' *Kappa* scores for each annotator in the main annotation task. As can be seen, the majority of the agreements fall between 0.7-0.8, indicating high agreement at the sentence-level. For the token-level, following TSD dataset [62], we computed the pairwise *Kappa* by using character offsets. Figure 2b shows the calculated scores. As can be seen, the majority of the agreements fall between 0.6-0.7. While the inter-annotator agreement is low compared to the sentence-level, it is comparable to similar token-level datasets such as TSD, where the mean pairwise *Kappa* was 0.55. Therefore, we believe that this agreement is reasonably high, given the highly subjective nature of the token-level offensive language identification task.

To decide on the gold label, we apply majority voting. For sentence-level offensive language identification, what more than two out of three annotators choose is selected as the gold label. Regarding offensive tokens, characters with more than two annotators annotate as offensive are provided as the ground truth.

---

[9]LightTag is available at https://www.lighttag.io/

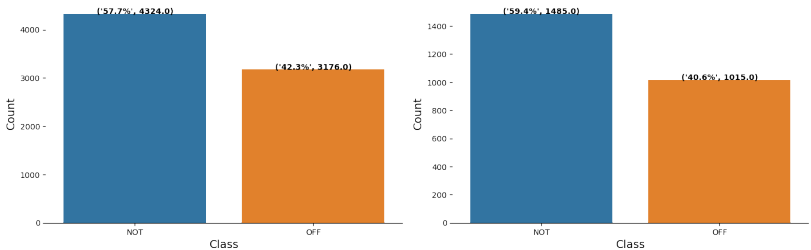

(a) Class distribution in training set   (b) Class distribution in testing set

**Figure 3**: Class distribution in *SOLD*.

## 4 *SOLD*: Sinhala Offensive Language Dataset

Table 1 shows several examples from the dataset along with English translations. The final dataset contains 10,000 tweets, of which 4191 tweets are annotated as offensive (41%). This is comparable to existing datasets for offensive language detection, where the number of offensive instances is much less than that of non-offensive instances. Furthermore, it is worth noticing that *SOLD* has a higher percentage of offensive instances compared to other low-resource datasets in the domain. For example, in RUHSOLD [54], only 24% of the Urdu tweets were considered offensive by the majority of the annotators, and in [49] only 19% of the Turkish tweets were labelled as offensive.

We divided the dataset into a training set and a testing set using a random split. The training set was used mainly to train the machine learning models, and the sole purpose of the testing set was to evaluate the trained machine learning models. Following the random split, 75% instances from the original dataset were assigned for the training set, and the rest of the instances were assigned for the testing test. The dataset is released as an open-access dataset in HuggingFace Datasets [107][10]. As can be seen in Figure 3, both training and testing sets have a similar distribution in the offensive and non-offensive classes.

We further analysed the length of the tweets as the length can be a limitation in attention-based neural networks [77]. As shown in Figure 4, most tweets have between 0-20 tokens. Both the offensive class and the non-offensive class follow a similar pattern in the token distribution. Since the number of tokens per tweet is relatively low, attention-based neural networks can be used to model the task without truncating the texts.

Table 2 shows the keywords used to collect SOLD and the percentage of offensive tweets for each keyword in training, testing and full datasets. As can be seen, these words are offensive based on the context, as the majority of the offensive percentages are between 30% - 50%. Therefore, a rule-based approach that depends on keywords will not perform successfully on this dataset.

---

[10]*SOLD* dataset can be downloaded from https://huggingface.co/datasets/sinhala-nlp/SOLD

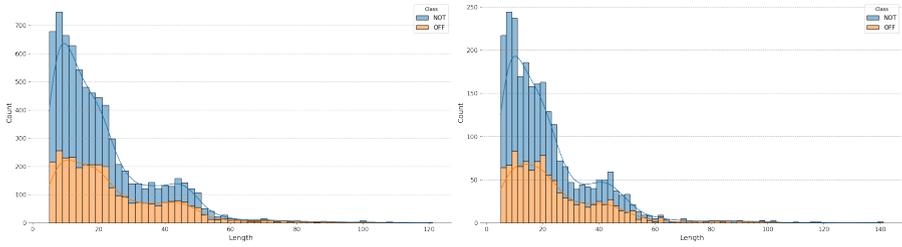

(a) Token frequency distribution in training set
(b) Token frequency distribution in testing set

**Figure 4**: Token frequency distribution in *SOLD*.

In the next section, we explore machine learning models that take context into account in detecting offensive language.

| Tweet | Class |
| --- | --- |
| @USER අඩෝ කෑම බීම හරියට තියෙනවද ? <br> (@USER Do you have enough foods and drinks?) | NOT |
| @USER @USER කොහොමවත් නෑ.. මම දන්නවනෙ ඒකිගෙ තොට තියෙන ආදරේ.. <br> (@USER @USER No. I know her love for you) | NOT |
| @USER @USER ඇය නලවනකොට බඩුව වගේ yo yo <br> (@USER @USER She is like a whore when she moves slowly yo yo) | OFF |
| @USER අපොයි මෙහෙමත් මොඩයෙක්. ජනදිපති අපෙක්ෂකයෙක් මීටවඩා බුද්දිමත් විදිහට කතාකලයුතුයි. <br> (@USER what a fool, a presidential candidate should speak intelligently than this) | OFF |

**Table 1**: Four tweets from the dataset, with their sentence level labels. Offensive tokens are highlisted in red. English translations are inside brackets

# 5 Experiments and Evaluation

The following sections will describe the experiments we conducted for sentence-level and token-level offensive language identification.

## 5.1 Sentence-level Offensive Language Detection

We consider sentence-level offensive language detection as a text classification task. We experimented with several ML text classifier models trained on the training set and evaluated them by predicting the labels for the held-out test set. As the label distribution is highly imbalanced, we evaluate and compare the performance of the different models using macro-averaged F1-score. We further report per-class Precision (P), Recall (R), F1-score (F1), and weighted average. The performance of the ML algorithms described below is shown in Table 3. All experiments were conducted using five different random seeds, and the mean value across these experiments is reported. Finally, we compare the performance of the models against the simple majority and minority class baselines.

| Index | Word (Translation) | Offensive % | | |
|---|---|---|---|---|
| | | Training | Testing | Total |
| 1 | තොපි (you) | 79.72 | 77.27 | 79.16 |
| 2 | ගොන් (foolish) | 75.74 | 72.82 | 75.00 |
| 3 | තෝ (you) | 72.00 | 72.45 | 72.12 |
| 4 | බැල්ලි (bitch) | 66.37 | 69.23 | 67.10 |
| 5 | තොට (to you) | 65.60 | 57.81 | 63.63 |
| 6 | හරකා (cow) | 60.97 | 55.88 | 59.87 |
| 7 | රෙද්ද (cloth/annoying thing) | 54.51 | 60.19 | 55.97 |
| 8 | යකෝ (you devil) | 49.90 | 53.65 | 50.78 |
| 9 | නාකි (old) | 50.00 | 54.34 | 51.06 |
| 10 | වල් (wild) | 49.87 | 48.02 | 49.37 |
| 11 | තමුසෙ (you) | 53.84 | 26.82 | 46.83 |
| 12 | පලයන් (go) | 47.64 | 38.29 | 45.62 |
| 13 | උඹ (you) | 46.19 | 44.05 | 45.62 |
| 14 | ගෑණි (woman) | 46.15 | 42.85 | 45.31 |
| 15 | තන් (breast) | 44.22 | 47.27 | 45.03 |
| 16 | ගල (stone/thigh) | 42.96 | 44.96 | 43.46 |
| 17 | හෙන (very/thunder) | 42.62 | 42.76 | 42.66 |
| 18 | සාප (curse) | 41.96 | 37.25 | 40.49 |
| 19 | අම්ම (mother) | 38.16 | 47.58 | 40.42 |
| 20 | බල්ලා (dog) | 37.88 | 48.93 | 40.38 |
| 21 | ගොනා (bull/prostitute) | 40.47 | 36.73 | 39.42 |
| 22 | පස්ස (back/ass) | 39.20 | 36.73 | 39.40 |
| 23 | පිස්සු (crazy) | 33.71 | 34.88 | 34.00 |
| 24 | පිස්සා (crazy person) | 34.90 | 30.55 | 33.80 |
| 25 | කරුම (curse) | 29.60 | 45.45 | 33.72 |
| 26 | බොල (curse) | 32.41 | 29.82 | 31.79 |
| 27 | ඒකි (she) | 27.50 | 39.43 | 31.16 |
| 28 | නඩ (big/fat) | 32.25 | 22.72 | 29.76 |
| 29 | පට්ට (extremely) | 25.00 | 23.77 | 24.67 |
| 30 | කැල්ල (girl) | 24.41 | 23.80 | 24.25 |
| 31 | කැත (ugly) | 23.27 | 16.66 | 21.89 |
| 32 | අමාරු (difficult) | 20.19 | 18.18 | 19.64 |
| 33 | කටට (to mouth) | 20.16 | 12.90 | 18.70 |
| 34 | අඩෝ (mate) | 16.00 | 7.69 | 13.66 |
| 35 | ඹහුඩු (wow) | 12.96 | 6.45 | 11.51 |

**Table 2**: The keywords used to collect *SOLD* and the percentage of offensive tweets for each keyword in training, testing and full datasets. Keywords are sorted from the offensive percentage in the full dataset.

#### *SVC*

Our simplest machine learning model is a linear Support Vector Classifier (SVC) trained on word unigrams. Before the emergence of neural networks, SVCs have achieved state-of-the-art results for many text classification tasks [108, 109] including offensive language identification [104, 110]. Even in the neural network era, SVCs produce an efficient and effective baseline.

#### *BiLSTM*

As the first embedding-based neural model, we experimented with a bidirectional Long Short-Term-Memory (BiLSTM) model, which we adopted from a pre-existing model for Greek offensive language identification [43]. The model consists of (i) an input embedding layer, (ii) two bidirectional LSTM layers, and (iii) two dense layers. The output of the final dense layer is ultimately passed through a softmax layer to produce the final prediction. The architecture diagram of the BiLSTM model is shown in Figure 5. Our BiLSTM layer has 64 units, while the first dense layer had 256 units.

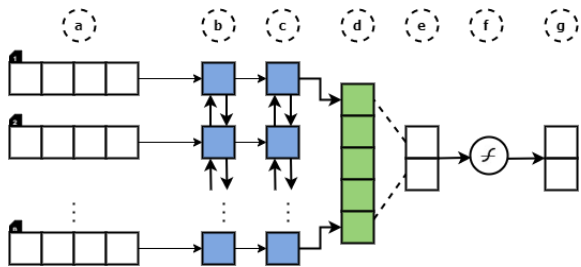

**Figure 5**: The BiLSTM model for sentence-level Sinhala offensive language identification. The labels are **(a)** input embeddings, **(b,c)** two BiLSTM layers, **(d, e)** fully-connected layers; **(f)** softmax activation, and **(g)** final probabilities [36]

#### *CNN*

We also experimented with a convolutional neural network (CNN), which we adopted from a pre-existing model for English sentiment classification [111]. The model consists of (i) an input embedding layer, (ii) 1 dimensional CNN layer (1DCNN), (iii) a max pooling layer and (iv) two dense layers. The output of the final dense layer is ultimately passed through a softmax layer to produce the final prediction.

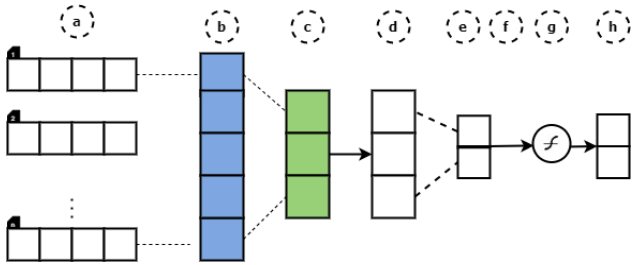

**Figure 6**: CNN model for sentence-level Sinhala offensive language identification. The labels are **(a)** input embeddings, **(b)** 1DCNN, **(c)** max pooling, **(d, e)** fully-connected layer; **(f)** with dropout, **(g)** softmax activation, and **(h)** final probabilities [36]

For the BiLSTM and CNN models presented above, we set three input channels for the input embedding layers: pre-trained Sinhala FastText embeddings[11] [112], Continuous Bag of Words Model for Sinhala[12] [113] as well as updatable embeddings learned by the model during training. For both models, we used the implementation provided in *OffensiveNN* Python library[13].

***Transformers***

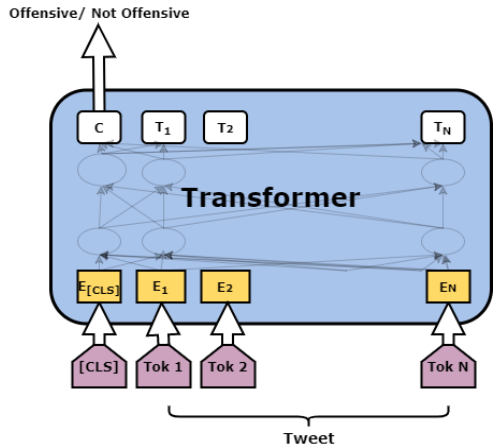

**Figure 7**: Transformer model for sentence-level Sinhala offensive language identification [7]

---

[11]Sinhala FastText embeddings are available on https://fasttext.cc/docs/en/crawl-vectors.html
[12]Sinhala word embeddings are available on https://github.com/nlpcuom/WEIntrinsicEvaluation
[13]OffensiveNN a pip package in https://pypi.org/project/offensivenn/

| Type | Model | OFF | | | NOT | | | Weighted | | | Macro F1 |
|---|---|---|---|---|---|---|---|---|---|---|---|
| | | P | R | F1 | P | R | F1 | P | R | F1 | |
| SVM | - | 0.63 | 0.44 | 0.52 | 0.68 | 0.82 | 0.75 | 0.66 | 0.67 | 0.65 | 0.63 |
| BiLSTM | CBOW | 0.70 | 0.75 | 0.72 | 0.82 | 0.78 | 0.80 | 0.77 | 0.77 | 0.77 | 0.76 |
| | fastText | 0.82 | 0.71 | 0.76 | 0.82 | 0.89 | 0.86 | 0.82 | 0.82 | 0.82 | 0.81 |
| | Self-learned | 0.66 | 0.34 | 0.45 | 0.66 | 0.88 | 0.76 | 0.66 | 0.66 | 0.63 | 0.60 |
| CNN | CBOW | 0.68 | 0.73 | 0.70 | 0.80 | 0.77 | 0.79 | 0.75 | 0.75 | 0.75 | 0.74 |
| | fastText | 0.82 | 0.73 | 0.77 | 0.83 | 0.89 | 0.86 | 0.83 | 0.83 | 0.82 | 0.82 |
| | Self-learned | 0.85 | 0.53 | 0.65 | 0.74 | 0.93 | 0.83 | 0.79 | 0.77 | 0.76 | 0.74 |
| Transformers | mBERT | 0.52 | 0.25 | 0.34 | 0.62 | 0.84 | 0.71 | 0.58 | 0.60 | 0.56 | 0.53 |
| | SinBERT | 0.80 | 0.75 | 0.77 | 0.83 | 0.87 | 0.85 | 0.82 | 0.82 | 0.82 | 0.81 |
| | XLM-R | 0.80 | 0.80 | 0.80 | 0.87 | 0.87 | 0.87 | 0.84 | 0.84 | 0.84 | **0.83** |
| | XLM-T | 0.79 | 0.79 | 0.79 | 0.85 | 0.86 | 0.85 | 0.83 | 0.83 | 0.83 | 0.82 |
| Baseline | All OFF | 0.41 | 1.00 | 0.58 | 0.00 | 0.00 | 0.00 | 0.16 | 0.41 | 0.23 | 0.29 |
| | All NOT | 0.00 | 0.00 | 0.00 | 0.59 | 1.00 | 0.75 | 0.35 | 0.59 | 0.44 | 0.37 |

**Table 3**: Results for offensive language detection sentence-level. **Type** refers to the machine learning algorithm used, and **Model** refers to the embedding model used. We report Precision (P), Recall (R), and F1 for each model/baseline on all classes (OFF, NOT) and weighted averages. Macro-F1 is also listed (best in bold).

Finally, we experimented with several pre-trained transformer models. With the introduction of BERT [76], transformer models have achieved state-of-the-art performance in many natural language processing tasks [76], including offensive language identification [7, 81]. From an input sentence, transformers compute a feature vector $\boldsymbol{h} \in \mathbb{R}^d$, upon which we build a classifier for the task. For this task, we implemented a softmax layer, i.e., the predicted probabilities are $\boldsymbol{y}^{(B)} = \text{softmax}(W\boldsymbol{h} + b)$, where $W \in \mathbb{R}^{k \times d}$ is the softmax weight matrix, and $k$ is the number of labels. In our experiments, we used three pre-trained transformer models available in HuggingFace model hub [114]; mBERT [76], SinBERT-large [115][14], xlm-roberta-large [83] (XLM-R) and XLM-T[116][15]. The implementation was adopted from the *DeepOffense* Python library[16]. The overall transformer architecture is available in Figure 7. For the transformer-based models, we employed a batch-size of 16, Adam optimiser with learning rate 2e−5, and a linear learning rate warm-up over 10% of the training data. During the training process, the parameters of the transformer model, as well as the parameters of the subsequent layers, were updated. The models were evaluated while training using an evaluation set that had one-fifth of the rows in training data. We performed early stopping if the evaluation loss did not improve over three evaluation steps. All the models were trained for three epochs.

---

[14]SinBERT is available at https://huggingface.co/NLPC-UOM/SinBERT-large
[15]XLM-T is trained over 198M tweets including Sinhala. The model is available at https://huggingface.co/cardiffnlp/twitter-xlm-roberta-base
[16]DeepOffense is available as a pip package in https://pypi.org/project/deepoffense/

As can be seen in Table 3, all models perform better than the majority baseline. As expected, neural models outperform the traditional machine learning model, SVM. From the experimented word embedding models, fastText [112] performed best, providing a 0.82 Macro F1 score with the CNN architecture, even outperforming language specific transformer models such as SinBERT [115]. The success of the CNN architecture in offensive language identification is similar to the previous research in English [104]. From the transformer models, mBERT [76] does not perform well because mBERT is not trained on Sinhala. The poor results of the mBERT suggest that advanced techniques are required when pre-trained language models are applied to unseen languages [117]. From all the models, XLM-R [83] performed best with a 0.83 Macro F1 score. This is closely followed by XLM-T [116] and CNN with fastText [112] having 0.82 Macro F1 scores.

## 5.2 Token-level Offensive Language Identification

We consider token-level offensive language detection as a token classification task. We experimented with several ML token classifier models trained on the training set and evaluated them by predicting the labels for the held-out test set. For the evaluation, we used the precision (P), Recall (R), and Macro F1 score of the offensive tokens. The performance of the ML algorithms described below is shown in Table 4. All experiments were conducted using five different random seeds, and the mean value across these experiments is reported.

### *BiLSTM*

As the first embedding-based neural model, we experimented with a BiLSTM model, which we adopted from a pre-existing model for English toxic spans detection task [86]. The model consists of (i) an input embedding layer, (ii) a bidirectional LSTM layer with 64 units, followed by (iii) a linear chain conditional random field (CRF) [118]. Similar to the previous experiments, we set three input channels for the input embedding layers: pre-trained Sinhala FastText embeddings [112], Continuous Bag of Words Model for Sinhala [113] as well as updatable embeddings learned by the model during training.

### *Transformers*

In token-level offensive language identification also, we experimented with several pre-trained transformer models. For a token classification task, transformer models add a linear layer that takes the last hidden state of the sequence as the input and produces a label for each token as the output. In this case, each token can have two labels; offensive and not offensive. In our experiments, we used the same three pre-trained transformer models we experimented with for sentence-level offensive language identification; mBERT [76], SinBERT-large [115], xlm-roberta-large [83] and XLM-T [116]. The implementation was adopted from the *MUDES* Python library[17]. The overall transformer architecture is available in Figure 8. For the transformer-based models, we employed a batch-size of 16, Adam optimiser with learning rate 2e−5, and a linear learning rate warm-up over 10% of the training data. During the training process, the parameters of the transformer model, as well as the parameters of the subsequent layers, were updated. The models were evaluated while training using an evaluation set that had one-fifth of the rows in training data. We performed early stopping if the evaluation loss did not improve over three evaluation steps. All the models were trained for three epochs.

---

[17]MUDES is available as a pip package in https://pypi.org/project/mudes/

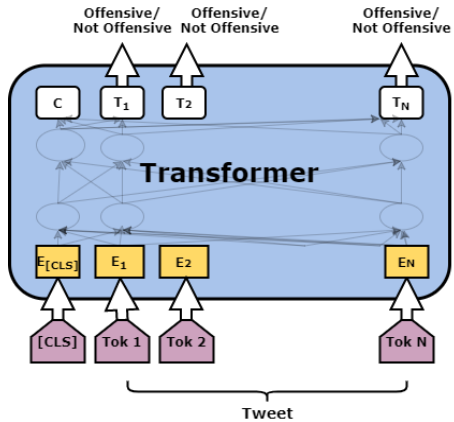

**Figure 8**: Transformer model for token-level Sinhala offensive language identification [93]

### Weakly Supervised Learning - Transformer+LIME

We utilised the binary classifiers that were trained to predict the offensive label of each post, and we employed LIME [58] at inference time to obtain offensive tokens [98, 99]. In LIME, new instances are generated by random sampling of the words that are present in the input. In other words, words are randomly left out from the input. The resulting new instances are then fed into the classifier, and a cloud of predicted probabilities is gathered. A linear model is then fitted, and coefficients for each token are the outputs of the LIME [58]. We obtain a sequence of binary decisions (offensive, not offensive) for the tokens of the post by using a probability threshold (tuned on one-fifth of the training data) applied to the LIME outputs for each token. We refer to this method as Transformer+LIME. This method requires only sentence-level offensive labels and does not require token-level annotations. Therefore, this is considered as a weakly supervised learning method [35]. We used the implementation provided in *lime* Python library[18].

---

[18]Lime is available as a pip package in https://pypi.org/project/lime/

| Type | Model | P | R | F1 |
|---|---|---|---|---|
| BiLSTM | CBOW | 0.44 | 0.73 | 0.58 |
|  | fastText | 0.48 | 0.74 | 0.60 |
|  | Self-learned | 0.40 | 0.70 | 0.55 |
| Transformers | SinBERT | 0.52 | 0.76 | 0.62 |
|  | XLM-R | 0.68 | 0.76 | **0.72** |
|  | XLM-T | 0.64 | 0.77 | 0.70 |
| Transformers + LIME | mBERT | 0.99 | 0.04 | 0.07 |
|  | SinBERT | 0.61 | 0.32 | 0.42 |
|  | XLM-R | 0.64 | 0.35 | 0.45 |
|  | XLM-T | 0.65 | 0.28 | 0.39 |
| Baseline | All OFF | 0.03 | 1.00 | 0.07 |

**Table 4**: Results for offensive language detection at token-level. **Type** refers to the machine learning algorithm used, and **Model** refers to the embedding model used. We report Precision (P), Recall (R), and F1 scores for the offensive tokens for each model/baseline (best in bold).

As can be seen in Table 4, all models perform better than the majority baseline. As expected, transformer models outperform the BiLSTM model. From the experimented word embedding models, fastText [112] performed best, similar to the sentence-level experiments. Additionally, we also experimented with mBERT [76]. However, the initial results showed that mBERT performs even worse than baselines. This shows that token-level offensive language identification is a difficult task for language models when the language is unseen in the pre-train process. From all the models, XLM-R [83] performed best with a 0.72 Macro F1 score similar to the sentence-level results. This is closely followed by XLM-T [116] having a 0.70 Macro F1 score. It is important to note that the transformer model trained specifically on Sinhala; SinBERT [115] did not perform well compared to the multilingual transformer models such as XLM-R [83].

In Table 4, we also show the weakly supervised learning results obtained with LIME [58]. Similar to the supervised models, XLM-R [83] performed best with a 0.45 Macro F1 score. Interestingly, XLM-T+LIME performs worse than SinBERT+LIME, despite the fact that the underlying XLM-T classifier is better (Macro F1 - 0.82) at sentence-level than the underlying SinBERT model (Macro F1 - 0.81). Overall, we can conclude that the weakly supervised models provided compatible results with the supervised models despite the fact that the latter is directly trained on offensive token annotations, whereas the former is trained with binary sentence-level annotations only.

With these results, we answer **RQ1:** *How do the state-of-the-art machine learning models perform in Sinhala offensive language identification at sentence-level and token-level?*. We showed that state-of-the-art machine learning models, such as XLM-R [83], perform well in identifying offence in both sentence and token levels. Furthermore, the results show that multilingual transformer models that support Sinhala, such as XLM-R [83] and [116], outperform language specific transformer model; SinBERT [115] in both sentence-level and token-level offensive language identification.

We also answer **RQ3.1:** *How to demonstrate explainability of the sentence-level offensive language identification using token-level annotations in Sinhala?* We employed LIME [58] on the transformer models trained at sentence-level and evaluated it using the token-level annotations in the test set. The results show that LIME based weakly supervised approach provides compatible results demonstrating the explainability of the sentence-level transformer models.

# 6 Transfer-learning Experiments

In a low resource language such as Sinhala, creating a large number of annotated instances can be a challenge due to the availability of qualified annotators. This is a huge limitation in improving the performance of machine learning models. The main goal of transfer learning experiments is to improve the performance of machine learning models in *SOLD* using an existing dataset without annotating more instances. As shown in Figure 9, in phase 1, we train a machine learning model on an existing dataset, and when we initialise the training process for *SOLD* in phase 2, we start with the saved weights from the phase 1. Since the majority of the existing datasets are from a different language, these experiments are usually referred to as *cross-lingual transfer learning.* As we discussed in Section 2, previous work has shown that a similar transfer learning approach can improve the results for Arabic, Greek, and Hindi [7, 32] at sentence-level offensive language identification.

In order to perform effective cross-lingual transfer learning, the underlying word representations in two languages need to be in the same vector space [7]. However, traditional word embedding models we used, such as FastText embeddings [112], and Continuous Bag of Words Model for Sinhala [113] are not in the same vector space with the word representations of English and other high-resource word embedding models [19]. Furthermore, initial experiments showed that the models based on FastText embeddings [112], and Continuous Bag of Words Model for Sinhala [113] do not improve with transfer learning. On the other hand, from the transformer models we experimented with, mBERT [119], XLM-R [83], and XLM-T [116], have shown cross-lingual properties. Therefore, we conduct the transfer learning experiments only with them.

This is the first time that cross-lingual transfer learning has been experimented with in Sinhala offensive language identification. Furthermore, cross-lingual transfer learning for token-level offensive language identification has not been explored before, which can be interesting for many languages.

---

[19]FastText [112] provides aligned word embeddings for 44 languages at https://fasttext.cc/docs/en/aligned-vectors.html. However, the language list does not include Sinhala

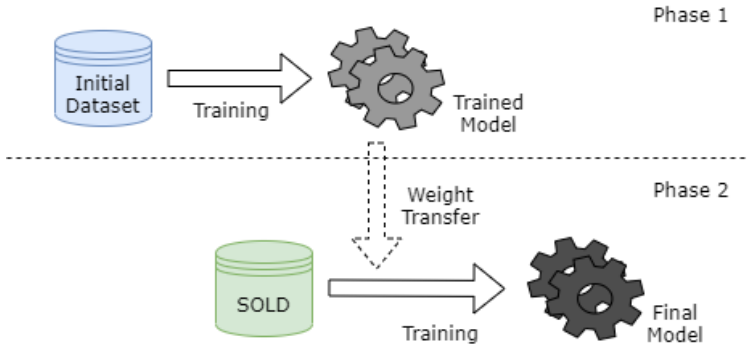

**Figure 9**: Transfer learning strategy

We used different resource-rich languages and datasets for sentence-level and token-level, which we describe in the following sections.

| Level | Dataset | Language | Inst. | Source | Labels |
|---|---|---|---|---|---|
| Sentence | OLID [104] | English | 14,100 | Twitter | offensive, non-offensive |
|  | HASOC [22] | Hindi | 8,000 | Twitter | hate offensive, non hate-offensive |
|  | CMCS [56] | Sinhala Code-mix | 10,000 | Twitter | hate-inducing, abusive, not offensive |
| Token | TSD [62] | English | 7939 | Civil Comments | toxic, not toxic |
|  | HateX [34] | English | 11,536 | Twitter, Gab | hate offensive, non hate-offensive |

**Table 5**: **Dataset** name, **Language**, number of instances (**Inst.**), **Source** and **Labels** in all initial resources used in transfer learning experiments. **Level** refers to whether we used the sentence-level or token-level annotations.

## 6.1 Sentence-level Offensive Language Detection

For the sentence-level, we used several resources as the initial dataset. As the first resource, we used OLID [104], arguably one of the most popular offensive language identification datasets in English. We specifically used the OLID [104] level A tweets, which is similar to the sentence-level of *SOLD*. Also, in order to perform transfer learning from a closely-related language to Sinhala, we utilised a Hindi dataset used in the HASOC 2020 shared task [22]. Hindi belongs to the Indo-Aryan language family, which is similar to Sinhala [120]. Furthermore, since both languages originated in the Indian subcontinent, they are also culturally closely related. In HASOC, instances are annotated at the sentence-level with *hate-offensive* and *non hate-offensive* [22]. We mapped the *hate-offensive* instances to our *offensive* class and *non hate-offensive* instances to our *not offensive* class, following our sentence-level annotation guidelines. We also used a recently released Sinhala code-mixed dataset (CMCS) [56] as the initial dataset in transfer-learning experiments. In CMCS, 10,000 instances have been annotated in three classes; Hate-Inducing, Abusive and Not offensive [56]. Before performing transfer learning, we mapped the Hate-Inducing and Abusive classes to a single offensive class following our definition of sentence-level offensive language labels. Mapping the offensive labels into a single offensive class has been a common approach in recent transfer learning research [7, 32]. These datasets are summarised in the first row in Table 5.

| Type | Model | Dataset | Weighted F1 (▲ %) | Macro F1 (▲ %) |
|---|---|---|---|---|
| Transformers | mBERT | Hindi<br>English<br>CMCS | 0.60 (+3.40)<br>0.59 (+2.96)<br>0.59 (+2.80) | 0.57 (+4.84)<br>0.57 (+4.37)<br>0.57 (+4.65) |
|  | XLM-R | Hindi<br>English<br>CMCS | **0.85 (+0.97)**<br>0.85 (+0.51)<br>0.84 (-0.32) | **0.84 (+1.02)**<br>0.84 (+0.56)<br>0.83 (-0.33) |
|  | XLM-T | Hindi<br>English<br>CMCS | 0.84 (+0.87)<br>0.84 (+1.53)<br>0.83 (+0.25) | 0.83 (+0.93)<br>0.84 (+1.56)<br>0.82 (+0.27) |
| Best | XLM-R | NA | 0.84 | 0.83 |

**Table 6**: Results for offensive language detection at sentence-level after transfer learning. **Type** refers to the machine learning algorithm used, **Model** refers to the embedding model used, and **Dataset** refers to the initial dataset that the model was trained on. We report weighted average F1 and macro F1 for each model (best in bold). With each result, we also report the difference of the same model with respect to non-transfer learning experiments in Table 3 as a percentage. The best result from Table 3 is shown in the last row.

All the datasets we used for transfer learning experiments at sentence-level contain Twitter data making them in-domain with respect to *SOLD*. However, since CMCS contains code-mixed texts, this will be the first time that transfer learning is experimented with between code-mixed Sinhala and Sinhala written in Sinhala script. As mentioned before, we conduct transfer learning experiments only with transformer models that have shown cross-lingual properties such as mBERT [119], XLM-R [83], and XLM-T [116].

Results of the transfer learning experiments at the sentence-level are shown in Table 6. As shown in the results, transfer learning improved results in all the experiments except when XLM-R [83] trained with CMCS [56]. The best result was given by XLM-R [83] when performing transfer learning with the Hindi dataset [22], which provided an improvement of more than 1% in Macro F1 when compared to the experiment without transfer learning. Furthermore, this is the best result achieved for the *SOLD* dataset at sentence-level.

However, there is no clear indication from these experiments that the closely related language, Hindi, has an impact on transfer learning performance. Hindi [22] provided a bigger improvement with XLM-R [83] while English [104] provided a bigger improvement with XLM-T [116]. We believe that the performance of transfer learning depends both on the initial dataset and underlying embeddings. It is important to note that transfer learning from the code-mixed Sinhala dataset, CMCS [56] provided fewer improvements compared to other datasets. In fact, CMCS [56] with XLM-R [83] reduced the results. We believe that the transformer models we experimented with have not seen code-mixed data in the training process, and therefore, they fail to align the embeddings between code-mixed Sinhala words and Sinhala words written in the Sinhala script. As a result, there is no advantage in using code-mixed data in transfer learning experiments.

## 6.2 Token-level Offensive Language Detection

For the token-level transfer learning experiments, we only used English datasets as token-level offensive labels are not available in other languages. We specifically used the HateXplain [34] token-level annotations and TSD [62] as the initial datasets. In HateXplain [34], instances are annotated as offensive or hateful at the sentence-level. The tokens have been annotated as to whether they contribute to the sentence-level label or not. The second dataset; TSD, was released as the official dataset in the Toxic Spans Detection task at SemEval 2021 (Task 5) [62]. In TSD [62], if a post is toxic, the tokens have been annotated on whether they make the text toxic or not. Similar to our sentence-level experiments, we mapped the tokens labelled as *toxic* to our *offensive* class and *not offensive* class otherwise. These datasets are summarised in the second row in Table 5.

HateXplain dataset we used for transfer learning experiments at token-level contains Twitter data [34] making them in-domain with respect to *SOLD*. However, the TSD dataset contains instances from an archive of the Civil Comments platform [62], a commenting plugin for independent news sites and therefore, making the dataset off-domain with respect to *SOLD*. This is the first time that transfer learning has experimented with token-level offensive language identification.

We also explore how transfer learning affects the explainability of sentence-level models. To do that, we performed LIME [58] on the sentence-level models that were trained following transfer learning in the previous section and evaluated them on the token-level labels. As far as we know, this is the first time that transfer learning in offensive language identification has been explored with LIME [58].

The results for the token-level transfer learning experiments based on TSD [62] and HateXplain [34] are reported in the "Transformers" row in Table 7. The results with sentece-level transfer learning and LIME are reported in the "Transformers + LIME" row in Table 7.

| Type | Model | Dataset | P (▲ %) | R (▲ %) | F1 (▲ %) |
|---|---|---|---|---|---|
| Transformers | XLM-R | HateX<br>TSD | 0.69 (+0.52)<br>0.70 (+0.82) | 0.76 (+0.23)<br>0.77 (+0.91) | 0.72 (+0.16)<br>**0.73 (+0.98)** |
|  | XLM-T | HateX<br>TSD | 0.64 (+0.12)<br>0.64 (+0.10) | 0.78 (+1.02)<br>0.77 (+0.42) | 0.70 (+0.21)<br>0.70 (+0.11) |
| Transformers + LIME | mBERT | Hindi<br>English<br>CMCS | 0.99 (-0.23)<br>0.99 (-0.12)<br>0.21 (-77.76) | 0.04 (+0.15)<br>0.04 (-0.28)<br>0.06 (+2.87) | 0.08 (+0.30)<br>0.07 (-0.06)<br>0.09 (+5.65) |
|  | XLM-R | Hindi<br>English<br>CMCS | 0.66 (+1.88)<br>0.70 (+5.67)<br>0.65 (-0.64) | 0.34 (+0.34)<br>0.28 (-6.43)<br>0.23 (-5.43) | 0.45 (+0.43)<br>0.40 (-5.04)<br>0.34 (-10.32) |
|  | XLM-T | Hindi<br>English<br>CMCS | 0.63 (-1.36)<br>0.63 (-1.48)<br>0.66 (-1.01) | 0.22 (-5.12)<br>0.30 (+2.02)<br>0.30 (+2.01) | 0.33 (-6.30)<br>0.41 (+1.52)<br>0.42 (+2.82) |
| Best | XLM-R | NA | 0.68 | 0.76 | 0.72 |

**Table 7**: Results for offensive language detection at token-level after transfer learning. **Type** refers to the machine learning algorithm used, **Model** refers to the embedding model used, and **Dataset** refers to the initial dataset that the model was trained on. We report Precision (P), Recall (R), and F1 for each model. With each result, we also report the difference of the same model with respect to non-transfer learning experiments in Table 4 as a percentage. The best result from 4 is shown in the last row.

As shown in Table 7, transfer learning improved results in all the supervised experiments. The best result was given by XLM-R [83] when performing transfer learning with the TSD [62], which provided an improvement of close to 1% in Macro F1 when compared to the experiment without transfer learning. Interestingly, TSD [62] is off-domain compared to SOLD, yet it provides the biggest improvement. Furthermore, this is the best result achieved for the *SOLD* dataset at token-level. Similar to sentence-level, there is no clear evidence of which initial dataset improves results mostly in transfer learning experiments, as it depends both on the initial dataset and underlying embeddings. Overall, we can conclude that transfer learning improves results in token-level offensive language identification for Sinhala.

While transfer learning improved results in supervised token-level offensive language identification models, transfer learning did not improve weakly-supervised models in the majority of the experiments. As shown in Table 7, the token-level results dropped in several weakly-supervised models after performing transfer learning. For example, in Table 6, we observed that XLM-R [83] with transfer learning performed from OLID [104] provided the strongest model for sentence-level offensive language identification. However, when the same model was employed with LIME [58] to predict token-level labels, the results dropped by 1% in Macro F1. This is an interesting observation, given that the underlying transformer model gets stronger with transfer learning, but it does not necessarily improve the explainability of the models.

With the findings in this section, we answer **RQ2.1:** *Do available resources from resource-rich languages combine with transfer-learning techniques aid the detection of offensive language in Sinhala at sentence-level and token-level?.* We performed transfer learning from different datasets and showed that transfer learning improves results in the majority of the experiments at sentence-level and all the experiments at token-level. The best results at both sentence-level and token-level that we have seen so far in *SOLD* were achieved after performing transfer learning in this section. These findings will be beneficial for many low-resource languages where the training data is scarce.

We also answer **RQ3.2** regarding explainability; *Does transfer-learning from resource-rich languages affect the explainability of the offensive language identification models?* We employed LIME [58] on sentence-level models that resulted after transfer learning to predict token-level offensive language in a weakly-supervised approach. The results indicate a performance drop in most of the models suggesting that transfer learning does not necessarily improve the explainability of the models. There is a large number of recent research that has used transfer learning to improve the results in sentence-level offensive language identification, [7, 51]; however, the researchers need to be aware of the fact that, transfer learning does not always improve the explainability. This finding will create a new direction in explainable ML research in offensive language identification.

# 7 Semi-supervised Data Augmentation

As we mentioned before, in a low resource language such as Sinhala, creating a large number of annotated instances is a challenge, and therefore, it is a major limitation in building ML models to detect offensive language. The second approach we propose to avoid this limitation is semi-supervised data augmentation which is also known as democratic co-learning [121]. This technique is used to create large datasets with noisy labels when provided with a set of diverse models trained in a supervised way. Semi-supervised data augmentation has improved results in multiple tasks, including English offensive language identification [37], sentiment analysis [122], and time series prediction [123].

In our work, we collected additional 145,000 Sinhala tweets using the same methods described in Section 3. Rather than labelling them manually, we used the ML models trained in Section 5 to label them. For each tweet in the unannotated dataset, each ML model in Section 5, predicts the confidence for the offensive class resulting in eleven confidence values for each tweet. We release this dataset; *SemiSOLD* as an open-access dataset in HuggingFace Datasets [107][20].

---

[20]*SemiSOLD* dataset can be downloaded from https://huggingface.co/datasets/sinhala-nlp/SemiSOLD

In the following sections, we detail how *SemiSOLD* was used in sentence-level and token-level experiments. As far as we know, this would be the first time that semi-supervised data augmentation is applied in Sinhala. Furthermore, semi-supervised data augmentation has not been explored before with explainable tokens, which can be interesting for many languages.

## 7.1 Sentence-level Offensive Language Detection

For the sentence-level, we used a filtering technique to filter the unannotated instances because the benefits of data augmentation can be hampered by noise in initial model predictions. We selected the three best sentence-level offensive language detection models from Section 5; XLM-R [83], XLM-T [116], and CNN with fastText [112]. For each instance in *SemiSOLD*, we calculated the standard deviation of the confidences of these three models for the positive class, which corresponds to the uncertainty of the models. We used different threshold values for model uncertainty to filter the data from *SemiSOLD*. For the labels, we compute an aggregated single prediction based on the average predicted by each of the above-mentioned models. If the average is greater than 0.5, we label the instance as *offensive*, and *not offensive* otherwise.

We used three threshold values; 0.05, 0.1 and 0.15. For each threshold value, we filter the instances in *SemiSOLD* and augment it to the training set of *SOLD*. We train the same ML models we experimented with in Section 5 on the augmented training set. We evaluated the results on the testing set of *SOLD*. The results are shown in Table 8.

As shown in the results, all the models benefitted from semi-supervised data augmentation. The best result was produced by XLM-R with a 0.1 threshold. We discover two key observations from the results. (1) Models only improve with 0.05 and 0.1. Despite having more instances in the 0.15 threshold, it does not improve the results in many ML models. This is mainly because the 0.15 threshold adds a large number of uncertain noisy instances to the training set, and ML models find it difficult to learn from these instances. (2) Smaller and lightweight models such as BiLSTM and CNN show notable improvements with data augmentation compared to large transformer models. This is similar to the previous experiments in data augmentation [37] where the results do not improve when the machine learning classifier is already strong. We can assume that the transformer models are already well trained for SOLD, and adding further instances to the training process would not improve the results for the transformer models.

With this finding, we answer **RQ2.2**: *Can semi-supervised data augmentation improve the results for Sinhala offensive language identification at sentence-level?* We showed that data augmentation could improve the results for ML models. However, it is important to find an optimal uncertainty threshold. As we demonstrated in the results, having too many noisy instances with a larger uncertainty threshold can lead to reduced performance in ML models.

The performance improvement of lightweight models can be an important research direction in knowledge distillation research. Knowledge distillation aims to extract knowledge from a top-performing large model into a smaller yet well performing model [124]. The smaller model is less demanding in terms of memory print and computing power and has a lower prediction latency encouraging green computing. Knowledge distillation has been explored in several NLP topics such as neural machine translation [125], language modelling [126], and translation quality estimation [127]. Therefore, the development of *SemiSOLD* can open new avenues for knowledge destabilisation in low resource offensive language identification.

| STD | Inst. | Type | Model | Weighted F1 (▲ %) | Macro F1 (▲ %) |
|---|---|---|---|---|---|
| 0.05 | 1819 | BiLSTM | CBOW | 0.78 (+1.02) | 0.77 (+1.19) |
| | | | fastText | 0.82 (+0.56) | 0.81 (+0.45) |
| | | | Self-learned | 0.66 (+3.34) | 0.64 (+4.03)) |
| | | CNN | CBOW | 0.75 (+0.45) | 0.75 (+0.23) |
| | | | fastText | 0.83 (+0.85) | 0.82 (+0.32) |
| | | | Self-learned | 0.78 (+0.98) | 0.77 (+0.76) |
| | | Transformers | mBERT | 0.60 (+3.76) | 0.56 (+3.54) |
| | | | SinBERT | 0.83 (+0.65) | 0.82 (+0.45) |
| | | | XLM-R | 0.84 (+0.15) | 0.83 (+0.11) |
| | | | XLM-T | 0.83 (+0.33) | 0.82 (+0.28) |
| 0.1 | 8474 | BiLSTM | CBOW | 0.80 (+2.42) | 0.79 (+2.78) |
| | | | fastText | 0.83 (+1.23) | 0.82 (+1.34) |
| | | | Self-learned | 0.70 (+7.25) | 0.68 (+8.09) |
| | | CNN | CBOW | 0.77 (+1.85) | 0.76 (+1.21) |
| | | | fastText | 0.84 (+1.43) | 0.83 (+1.35) |
| | | | Self-learned | 0.79 (+1.56) | 0.78 (+1.76) |
| | | Transformers | mBERT | 0.63 (+7.09) | 0.59 (+6.21) |
| | | | SinBERT | 0.84 (+1.11) | 0.83 (+1.03) |
| | | | XLM-R | **0.85 (+0.72)** | **0.84** (+0.63) |
| | | | XLM-T | 0.84 (+0.86) | 0.83 (+0.92) |
| 0.15 | 47746 | BiLSTM | CBOW | 0.75 (-1.32) | 0.73 (-1.86) |
| | | | fastText | 0.80 (-1.43) | 0.79 (-1.56) |
| | | | Self-learned | 0.65 (+2.43) | 0.61 (+1.82) |
| | | CNN | CBOW | 0.75 (-0.32) | 0.74 (-0.54) |
| | | | fastText | 0.80 (-2.54) | 0.79 (-2.89) |
| | | | Self-learned | 0.78 (+0.68) | 0.77 (+0.82) |
| | | Transformers | mBERT | 0.61 (+5.65) | 0.58 (+4.78) |
| | | | SinBERT | 0.81 (-1.28) | 0.80 (-1.93) |
| | | | XLM-R | 0.82 (-1.84) | 0.81 (-2.08) |
| | | | XLM-T | 0.82 (-1.68) | 0.80 (-1.95) |
| | | Best | XLM-R | 0.84 | 0.83 |

**Table 8**: Results for offensive language detection at sentence-level after data augmentation. **STD** shows the uncertainty threshold and **Inst.** is the number of total unlabelled instances augmented. **Type** refers to the machine learning algorithm used, and **Model** refers to the embedding model used. We report weighted average F1 and macro F1 for each model (best in bold). With each result, we also report the difference of the same model with respect to non-transfer learning experiments in Table 3 as a percentage. The best result from Table 3 is shown in the last row.

| STD | Inst. | Type | Model | P (▲ %) | R (▲ %) | F1 (▲ %) |
|---|---|---|---|---|---|---|
| 0.05 | 1819 | Transformers + LIME | mBERT | 0.99 (+0.02) | 0.04 (+0.04) | 0.07 (+0.10) |
| | | | SinBERT | 0.62 (+1.45) | 0.33 (+1.25) | 0.43 (+1.85) |
| | | | XLM-R | 0.65 (+1.24) | 0.36 (+1.68) | 0.46 (+2.21) |
| | | | XLM-T | 0.66 (+1.18) | 0.30 (+1.82) | 0.40 (+0.80) |
| 0.1 | 8474 | | mBERT | 0.98 (-1.08) | 0.06 (+2.23) | 0.09 (+1.82) |
| | | | SinBERT | 0.64 (+3.56) | 0.35 (+3.78) | 0.45 (+3.56) |
| | | | XLM-R | 0.66 (+1.15) | 0.38 (+3.21) | **0.48 (+3.79)** |
| | | | XLM-T | 0.65 (+0.21) | 0.32 (+4.32) | 0.44 (+5.34) |
| 0.15 | 47746 | | mBERT | 0.99 (+0.04) | 0.05 (+1.21) | 0.08 (+1.08) |
| | | | SinBERT | 0.62 (+1.23) | 0.28 (-3.85) | 0.39 (-2.97) |
| | | | XLM-R | 0.64 (+0.56) | 0.33 (-3.56) | 0.42 (-2.99) |
| | | | XLM-T | 0.65 (-0.43) | 0.24 (-4.76) | 0.34 (-5.68) |
| | | Best | XLM-R | 0.68 | 0.76 | 0.72 |

**Table 9**: Results for offensive language detection at token-level after data augmentation. **STD** shows the uncertainty threshold and **Inst.** is the number of total unlabelled instances augmented. **Type** refers to the machine learning algorithm used, and **Model** refers to the embedding model used. We report Precision (P), Recall (R), and F1 for each model/baseline (best in bold).

### 7.2 Token-level Offensive Language Detection

For the token classification tasks, the semi-supervised data augmentation technique we used with democratic co-learning and model uncertainty does not readily apply. While sentence-level seeks to minimise the divergence between the outputs of different models, for token classification, the number of label combinations grows exponentially with respect to the sequence length. Extracting model knowledge as if each combination is a different label category would be largely inefficient [128].

Considering this, we do not train supervised token-level models on the augmented data. Rather than that, we used the sentence-level models trained on augmented data in Section 7.1 to predict token-level labels using LIME, as we discussed in previous sections. The results are shown in Table 9.

As can be seen in Table 9, data augmentation improved the results of weakly supervised models in token-level offensive language detection. The best F1 score for "Transformers + LIME" was achieved with XLM-R [83] and 0.1 model uncertainty. Similar to the previous section, we notice a drop in the results with 0.15 model uncertainty. This is mainly because the noisy instances in the 0.15 threshold have weakened the sentence-level models, as we saw in Table 8 and therefore, they do not provide better results with LIME. Overall, 0.48 is the best result got for "Transformers + LIME" with *SOLD*.

With this finding, we answer **RQ3.3**: *Can semi-supervised data augmentation improve the explainability of the sentence-level models?*. As we experimented with LIME and transformers, we showed that data augmentation could improve explainability. However, it is important not to follow a greedy approach with data augmentation and only augment less noisy instances. Adding more noisy instances can lead to a weakened sentence-level model, which could impact the explainability.

Several large offensive language datasets with sentence-level annotations are publicly available for many languages. For the languages that do not have large offensive language datasets, it is straightforward to collect more data following a similar methodology we used to collect *SemiSOLD*. As we showed, the weakly supervised offensive token detector, "Transformers + LIME", can, in principle, perform even better if the underlying binary classifier is trained on a larger dataset. Therefore, this finding can be a huge step towards explainable offensive language detection in many languages.

# 8 Conclusion and Future Work

In this paper, we presented a comprehensive evaluation of Sinhala offensive language identification along with two new resources: *SOLD* and *SemiSOLD*. *SOLD* contains 10,500 tweets annotated at sentence-level and token-level, making it the largest manually annotated Sinhala offensive language dataset to date. *SemiSOLD* is a larger dataset of more than 145,000 instances annotated with semi-supervised methods. Both these results open exciting new avenues for research on Sinhala and other low-resource languages.

Our results show that state-of-the-art ML models can be used to identify Sinhala offensive language at sentence and token-level (answering **RQ1**). With respect to **RQ2** addressing data scarcity in low-resource languages, we report that (1) transfer learning techniques from both English and Hindi result in performance improvement for Marathi in sentence-level and token-level offensive language detection (answering **RQ2.1**) (2) the use of the larger dataset *SemiSOLD* combined with *SOLD* results in performance improvement for sentence-level offensive language identification, particularly for lightweight models such as BiLSTM and CNN (answering **RQ2.2**). With respect to **RQ3** addressing explainability, we report that (1) transformer models trained on sentence-level combined with LIME can be used to predict offensive tokens demonstrating their explainability (answering **RQ3.1** (2) sentence-level transfer learning from resource-rich languages do not necessarily improve explainability despite having a strong sentence-level model (answering **RQ3.2** (3) semi-supervised data augmentation on sentence-level can improve the explainability (answering **RQ3.3**. We believe that these results shed light on offensive language identification applied to Sinhala and other low-resource languages as well.

In future work, we would like to extend SOLD's annotation to type and target annotations in offensive posts. This would allow us to identify common targets in Sinhala offensive social media posts and prevent targeted offence towards certain individuals and groups. We would also like to extend the dataset to other platforms, such as YouTube comments and news media comments. Finally, we would like to use the knowledge and data obtained from our work on Sinhala and expand it to closely-related Indo-Aryan languages to Sinhala, such as Dhivehi.

## Statements and Declarations

- **Funding** - None
- **Conflict of interest/Competing interests** - The authors declare that they have no conflict of interest.
- **Availability of data** - Data is available at https://huggingface.co/sinhala-nlp.
- **Availability of code** - Code is available at https://github.com/Sinhala-NLP/SOLD.
- **Authors' contributions** -
    - Tharindu Ranasinghe - Problem formulation, Dataset formulation, Conducting experiments, Writing
    - Isuri Anuradha - Dataset formulation, Conducting experiments
    - Damith Premasiri - Dataset formulation, Conducting experiments
    - Kanishka Silva - Dataset formulation, Conducting experiments
    - Lasitha Uyangodage - Dataset formulation, Conducting experiments
    - Hansi Hettiarachchi - Dataset formulation, Writing
    - Marcos Zampieri - Problem formulation, Writing
- **Acknowledgements** - We want to acknowledge Janitha Hapuarachchi, Sachith Suraweera, Chandika Udaya Kumara and Ridmi Randima, the team of volunteer annotators that provided their free time and efforts to help us produce *SOLD*.

## References

[1] Hamm, M.P., Newton, A.S., Chisholm, A., Shulhan, J., Milne, A., Sundar, P., Ennis, H., Scott, S.D., Hartling, L.: Prevalence and Effect of Cyberbullying on Children and Young People: A Scoping Review of Social Media Studies. JAMA Pediatrics **169**(8), 770–777 (2015). https://doi.org/10.1001/jamapediatrics.2015.0944

[2] López-Meneses, E., Vázquez-Cano, E., González-Zamar, M.-D., Abad-Segura, E.: Socioeconomic effects in cyberbullying: Global research

trends in the educational context. International Journal of Environmental Research and Public Health **17**(12) (2020). https://doi.org/10.3390/ijerph17124369

[3] Bonanno, R.A., Hymel, S.: Cyber bullying and internalizing difficulties: Above and beyond the impact of traditional forms of bullying. Journal of Youth and Adolescence **42**(5), 685–697 (2013). https://doi.org/10.1007/s10964-013-9937-1

[4] Edwards, L., Kontostathis, A., Fisher, C.: Cyberbullying, race/ethnicity and mental health outcomes: A review of the literature. Media and Communication **4**(3), 71–78 (2016). https://doi.org/10.17645/mac.v4i3.525

[5] Vidgen, B., Nguyen, D., Margetts, H., Rossini, P., Tromble, R.: Introducing CAD: the contextual abuse dataset. In: Proceedings of the 2021 Conference of the North American Chapter of the Association for Computational Linguistics: Human Language Technologies, pp. 2289–2303. Association for Computational Linguistics, Online (2021). https://doi.org/10.18653/v1/2021.naacl-main.182. https://aclanthology.org/2021.naacl-main.182

[6] Pamungkas, E.W., Basile, V., Patti, V.: Towards multidomain and multilingual abusive language detection: a survey. Personal and Ubiquitous Computing (2021). https://doi.org/10.1007/s00779-021-01609-1

[7] Ranasinghe, T., Zampieri, M.: Multilingual offensive language identification with cross-lingual embeddings. In: Proceedings of the 2020 Conference on Empirical Methods in Natural Language Processing (EMNLP), pp. 5838–5844. Association for Computational Linguistics, Online (2020). https://doi.org/10.18653/v1/2020.emnlp-main.470. https://aclanthology.org/2020.emnlp-main.470

[8] Cercas Curry, A., Abercrombie, G., Rieser, V.: ConvAbuse: Data, analysis, and benchmarks for nuanced abuse detection in conversational AI. In: Proceedings of the 2021 Conference on Empirical Methods in Natural Language Processing, pp. 7388–7403. Association for Computational Linguistics, Online and Punta Cana, Dominican Republic (2021). https://doi.org/10.18653/v1/2021.emnlp-main.587. https://aclanthology.org/2021.emnlp-main.587

[9] Pamungkas, E.W., Basile, V., Patti, V.: Do you really want to hurt me? predicting abusive swearing in social media. In: Proceedings of the 12th Language Resources and Evaluation Conference, pp. 6237–6246. European Language Resources Association, Marseille, France (2020). https://aclanthology.org/2020.lrec-1.765

[10] Wulczyn, E., Thain, N., Dixon, L.: Ex machina: Personal attacks seen at scale. In: Proceedings of the 26th International Conference on World Wide Web. WWW '17, pp. 1391–1399. International World Wide Web Conferences Steering Committee, Republic and Canton of Geneva, CHE (2017). https://doi.org/10.1145/3038912.3052591. https://doi.org/10.1145/3038912.3052591

[11] Kumar, R., Reganti, A.N., Bhatia, A., Maheshwari, T.: Aggression-annotated corpus of Hindi-English code-mixed data. In: Proceedings of the Eleventh International Conference on Language Resources and Evaluation (LREC 2018). European Language Resources Association (ELRA), Miyazaki, Japan (2018). https://aclanthology.org/L18-1226

[12] Sprugnoli, R., Menini, S., Tonelli, S., Oncini, F., Piras, E.: Creating a WhatsApp dataset to study pre-teen cyberbullying. In: Proceedings of the 2nd Workshop on Abusive Language Online (ALW2), pp. 51–59. Association for Computational Linguistics, Brussels, Belgium (2018). https://doi.org/10.18653/v1/W18-5107. https://aclanthology.org/W18-5107

[13] Rosa, H., Pereira, N., Ribeiro, R., Ferreira, P.C., Carvalho, J.P., Oliveira, S., Coheur, L., Paulino, P., Veiga Simão, A.M., Trancoso, I.: Automatic cyberbullying detection: A systematic review. Computers in Human Behavior **93**, 333–345 (2019). https://doi.org/10.1016/j.chb.2018.12.021

[14] Wijesiriwardene, T., Inan, H., Kursuncu, U., Gaur, M., Shalin, V.L., Thirunarayan, K., Sheth, A., Arpinar, I.B.: Alone: A dataset for toxic behavior among adolescents on twitter. In: Social Informatics, pp. 427–439. Springer, Cham (2020)

[15] Borkan, D., Dixon, L., Sorensen, J., Thain, N., Vasserman, L.: Nuanced metrics for measuring unintended bias with real data for text classification. In: Companion Proceedings of The 2019 World Wide Web Conference. WWW '19, pp. 491–500. Association for Computing Machinery, New York, NY, USA (2019). https://doi.org/10.1145/3308560.3317593. https://doi.org/10.1145/3308560.3317593

[16] Fanton, M., Bonaldi, H., Tekiroğlu, S.S., Guerini, M.: Human-in-the-loop for data collection: a multi-target counter narrative dataset to fight online hate speech. In: Proceedings of the 59th Annual Meeting of the Association for Computational Linguistics and the 11th International Joint Conference on Natural Language Processing (Volume 1: Long Papers), pp. 3226–3240. Association for Computational Linguistics, Online (2021). https://doi.org/10.18653/v1/2021.acl-long.250. https://aclanthology.org/2021.acl-long.250

[17] Pitsilis, G.K., Ramampiaro, H., Langseth, H.: Effective hate-speech detection in twitter data using recurrent neural networks. Applied Intelligence **48**(12), 4730–4742 (2018). https://doi.org/10.1007/s10489-018-1242-y

[18] Zampieri, M., Malmasi, S., Nakov, P., Rosenthal, S., Farra, N., Kumar, R.: SemEval-2019 task 6: Identifying and categorizing offensive language in social media (OffensEval). In: Proceedings of the 13th International Workshop on Semantic Evaluation, pp. 75–86. Association for Computational Linguistics, Minneapolis, Minnesota, USA (2019). https://doi.org/10.18653/v1/S19-2010. https://aclanthology.org/S19-2010

[19] Zampieri, M., Nakov, P., Rosenthal, S., Atanasova, P., Karadzhov, G., Mubarak, H., Derczynski, L., Pitenis, Z., Çöltekin, Ç.: SemEval-2020 task 12: Multilingual offensive language identification in social media (OffensEval 2020). In: Proceedings of the Fourteenth Workshop on Semantic Evaluation, pp. 1425–1447. International Committee for Computational Linguistics, Barcelona (online) (2020). https://doi.org/10.18653/v1/2020.semeval-1.188. https://aclanthology.org/2020.semeval-1.188

[20] Kumar, R., Ojha, A.K., Malmasi, S., Zampieri, M.: Benchmarking aggression identification in social media. In: Proceedings of the First Workshop on Trolling, Aggression and Cyberbullying (TRAC-2018), pp. 1–11. Association for Computational Linguistics, Santa Fe, New Mexico, USA (2018). https://aclanthology.org/W18-4401

[21] Kumar, R., Ojha, A.K., Malmasi, S., Zampieri, M.: Evaluating aggression identification in social media. In: Proceedings of the Second Workshop on Trolling, Aggression and Cyberbullying, pp. 1–5. European Language Resources Association (ELRA), Marseille, France (2020). https://aclanthology.org/2020.trac-1.1

[22] Mandl, T., Modha, S., Kumar M, A., Chakravarthi, B.R.: Overview of the hasoc track at fire 2020: Hate speech and offensive language identification in tamil, malayalam, hindi, english and german. In: Forum for Information Retrieval Evaluation. FIRE 2020, pp. 29–32. Association for Computing Machinery, New York, NY, USA (2020). https://doi.org/10.1145/3441501.3441517. https://doi.org/10.1145/3441501.3441517

[23] Modha, S., Mandl, T., Shahi, G.K., Madhu, H., Satapara, S., Ranasinghe, T., Zampieri, M.: Overview of the hasoc subtrack at fire 2021: Hate speech and offensive content identification in english and indo-aryan languages and conversational hate speech. In: Forum for Information Retrieval Evaluation. FIRE 2021, pp. 1–3. Association for Computing Machinery, New York, NY, USA (2021). https://doi.org/10.

1145/3503162.3503176. https://doi.org/10.1145/3503162.3503176

[24] Basile, V., Bosco, C., Fersini, E., Nozza, D., Patti, V., Rangel Pardo, F.M., Rosso, P., Sanguinetti, M.: SemEval-2019 task 5: Multilingual detection of hate speech against immigrants and women in Twitter. In: Proceedings of the 13th International Workshop on Semantic Evaluation, pp. 54–63. Association for Computational Linguistics, Minneapolis, Minnesota, USA (2019). https://doi.org/10.18653/v1/S19-2007. https://aclanthology.org/S19-2007

[25] Caselli, T., Basile, V., Mitrović, J., Kartoziya, I., Granitzer, M.: I feel offended, don't be abusive! implicit/explicit messages in offensive and abusive language. In: Proceedings of the 12th Language Resources and Evaluation Conference, pp. 6193–6202. European Language Resources Association, Marseille, France (2020). https://aclanthology.org/2020.lrec-1.760

[26] Davidson, T., Warmsley, D., Macy, M.W., Weber, I.: Automated hate speech detection and the problem of offensive language. In: Proceedings of the International AAAI Conference on Web and Social Media (ICWSM), pp. 512–515 (2017)

[27] Mubarak, H., Rashed, A., Darwish, K., Samih, Y., Abdelali, A.: Arabic offensive language on Twitter: Analysis and experiments. In: Proceedings of the Sixth Arabic Natural Language Processing Workshop, pp. 126–135. Association for Computational Linguistics, Kyiv, Ukraine (Virtual) (2021). https://aclanthology.org/2021.wanlp-1.13

[28] Mubarak, H., Darwish, K., Magdy, W., Elsayed, T., Al-Khalifa, H.: Overview of OSACT4 Arabic offensive language detection shared task. In: Proceedings of the 4th Workshop on Open-Source Arabic Corpora and Processing Tools, with a Shared Task on Offensive Language Detection, pp. 48–52. European Language Resource Association, Marseille, France (2020). https://aclanthology.org/2020.osact-1.7

[29] Sigurbergsson, G.I., Derczynski, L.: Offensive language and hate speech detection for Danish. In: Proceedings of the 12th Language Resources and Evaluation Conference, pp. 3498–3508. European Language Resources Association, Marseille, France (2020). https://aclanthology.org/2020.lrec-1.430

[30] Caselli, T., Schelhaas, A., Weultjes, M., Leistra, F., van der Veen, H., Timmerman, G., Nissim, M.: DALC: the Dutch abusive language corpus. In: Proceedings of the 5th Workshop on Online Abuse and Harms (WOAH 2021), pp. 54–66. Association for Computational Linguistics, Online (2021). https://doi.org/10.18653/v1/2021.woah-1.6. https://aclanthology.org/2021.woah-1.6

[31] Chiril, P., Benamara Zitoune, F., Moriceau, V., Coulomb-Gully, M., Kumar, A.: Multilingual and multitarget hate speech detection in tweets. In: Actes de la Conférence sur Le Traitement Automatique des Langues Naturelles (TALN) PFIA 2019. Volume II : Articles Courts, pp. 351–360. ATALA, Toulouse, France (2019). https://aclanthology.org/2019.jeptalnrecital-court.21

[32] Ranasinghe, T., Zampieri, M.: Multilingual offensive language identification for low-resource languages. ACM Trans. Asian Low-Resour. Lang. Inf. Process. **21**(1) (2021). https://doi.org/10.1145/3457610

[33] Hase, P., Bansal, M.: Evaluating explainable AI: Which algorithmic explanations help users predict model behavior? In: Proceedings of the 58th Annual Meeting of the Association for Computational Linguistics, pp. 5540–5552. Association for Computational Linguistics, Online (2020). https://doi.org/10.18653/v1/2020.acl-main.491. https://aclanthology.org/2020.acl-main.491

[34] Mathew, B., Saha, P., Yimam, S.M., Biemann, C., Goyal, P., Mukherjee, A.: Hatexplain: A benchmark dataset for explainable hate speech detection. Proceedings of the AAAI Conference on Artificial Intelligence **35**(17), 14867–14875 (2021)

[35] Pavlopoulos, J., Laugier, L., Xenos, A., Sorensen, J., Androutsopoulos, I.: From the detection of toxic spans in online discussions to the analysis of toxic-to-civil transfer. In: Proceedings of the 60th Annual Meeting of the Association for Computational Linguistics (Volume 1: Long Papers), pp. 3721–3734. Association for Computational Linguistics, Dublin, Ireland (2022). https://doi.org/10.18653/v1/2022.acl-long.259. https://aclanthology.org/2022.acl-long.259

[36] Zampieri, M., Ranasinghe, T., Chaudhari, M., Gaikwad, S., Krishna, P., Nene, M., Paygude, S.: Predicting the type and target of offensive social media posts in marathi. Social Network Analysis and Mining **12**(1), 77 (2022). https://doi.org/10.1007/s13278-022-00906-8

[37] Rosenthal, S., Atanasova, P., Karadzhov, G., Zampieri, M., Nakov, P.: SOLID: A large-scale semi-supervised dataset for offensive language identification. In: Findings of the Association for Computational Linguistics: ACL-IJCNLP 2021, pp. 915–928. Association for Computational Linguistics, Online (2021). https://doi.org/10.18653/v1/2021.findings-acl.80. https://aclanthology.org/2021.findings-acl.80

[38] Ranasinghe, T., Zampieri, M., Hettiarachchi, H.: BRUMS at HASOC 2019: Deep Learning Models for Multilingual Hate Speech and Offensive Language Identification. In: Forum for Information Retrieval Evaluation, pp. 199–207 (2019)

[39] Liu, P., Li, W., Zou, L.: NULI at SemEval-2019 task 6: Transfer learning for offensive language detection using bidirectional transformers. In: Proceedings of the 13th International Workshop on Semantic Evaluation, pp. 87–91. Association for Computational Linguistics, Minneapolis, Minnesota, USA (2019). https://doi.org/10.18653/v1/S19-2011. https://aclanthology.org/S19-2011

[40] Yao, M., Chelmis, C., Zois, D.-S.: Cyberbullying ends here: Towards robust detection of cyberbullying in social media. In: The World Wide Web Conference. WWW '19, pp. 3427–3433. Association for Computing Machinery, New York, NY, USA (2019). https://doi.org/10.1145/3308558.3313462. https://doi.org/10.1145/3308558.3313462

[41] Ridenhour, M., Bagavathi, A., Raisi, E., Krishnan, S.: Detecting online hate speech: Approaches using weak supervision and network embedding models. In: Thomson, R., Bisgin, H., Dancy, C., Hyder, A., Hussain, M. (eds.) Social, Cultural, and Behavioral Modeling, pp. 202–212. Springer, Cham (2020)

[42] Assenmacher, D., Niemann, M., Müller, K., Seiler, M., Riehle, D.M., Trautmann, H.: Rp-mod & rp-crowd: Moderator- and crowd-annotated german news comment datasets. In: Thirty-fifth Conference on Neural Information Processing Systems Datasets and Benchmarks Track (Round 2) (2021). https://openreview.net/forum?id=NfTU-wN8Uo

[43] Pitenis, Z., Zampieri, M., Ranasinghe, T.: Offensive language identification in Greek. In: Proceedings of the 12th Language Resources and Evaluation Conference, pp. 5113–5119. European Language Resources Association, Marseille, France (2020). https://aclanthology.org/2020.lrec-1.629

[44] Sanguinetti, M., Poletto, F., Bosco, C., Patti, V., Stranisci, M.: An Italian Twitter corpus of hate speech against immigrants. In: Proceedings of the Eleventh International Conference on Language Resources and Evaluation (LREC 2018). European Language Resources Association (ELRA), Miyazaki, Japan (2018). https://aclanthology.org/L18-1443

[45] Fortuna, P., Rocha da Silva, J., Soler-Company, J., Wanner, L., Nunes, S.: A hierarchically-labeled Portuguese hate speech dataset. In: Proceedings of the Third Workshop on Abusive Language Online, pp. 94–104. Association for Computational Linguistics, Florence, Italy (2019). https://doi.org/10.18653/v1/W19-3510. https://aclanthology.org/W19-3510

[46] Moon, J., Cho, W.I., Lee, J.: BEEP! Korean corpus of online news comments for toxic speech detection. In: Proceedings of the Eighth International Workshop on Natural Language Processing

for Social Media, pp. 25–31. Association for Computational Linguistics, Online (2020). https://doi.org/10.18653/v1/2020.socialnlp-1.4. https://aclanthology.org/2020.socialnlp-1.4

[47] Ljubešić, N., Erjavec, T., Fišer, D.: Datasets of Slovene and Croatian moderated news comments. In: Proceedings of the 2nd Workshop on Abusive Language Online (ALW2), pp. 124–131. Association for Computational Linguistics, Brussels, Belgium (2018). https://doi.org/10.18653/v1/W18-5116. https://aclanthology.org/W18-5116

[48] Plaza-del-Arco, F.M., Montejo-Ráez, A., Ureña-López, L.A., Martín-Valdivia, M.-T.: OffendES: A new corpus in Spanish for offensive language research. In: Proceedings of the International Conference on Recent Advances in Natural Language Processing (RANLP 2021), pp. 1096–1108. INCOMA Ltd., Held Online (2021). https://aclanthology.org/2021.ranlp-1.123

[49] Çöltekin, Ç.: A corpus of Turkish offensive language on social media. In: Proceedings of the 12th Language Resources and Evaluation Conference, pp. 6174–6184. European Language Resources Association, Marseille, France (2020). https://aclanthology.org/2020.lrec-1.758

[50] Romim, N., Ahmed, M., Talukder, H., Saiful Islam, M.: Hate speech detection in the bengali language: A dataset and its baseline evaluation. In: Uddin, M.S., Bansal, J.C. (eds.) Proceedings of International Joint Conference on Advances in Computational Intelligence, pp. 457–468. Springer, Singapore (2021)

[51] Gaikwad, S.S., Ranasinghe, T., Zampieri, M., Homan, C.: Cross-lingual offensive language identification for low resource languages: The case of Marathi. In: Proceedings of the International Conference on Recent Advances in Natural Language Processing (RANLP 2021), pp. 437–443. INCOMA Ltd., Held Online (2021). https://aclanthology.org/2021.ranlp-1.50

[52] Niraula, N.B., Dulal, S., Koirala, D.: Offensive language detection in Nepali social media. In: Proceedings of the 5th Workshop on Online Abuse and Harms (WOAH 2021), pp. 67–75. Association for Computational Linguistics, Online (2021). https://doi.org/10.18653/v1/2021.woah-1.7. https://aclanthology.org/2021.woah-1.7

[53] Abainia, K., Kara, K., Hamouni, T.: A new corpus and lexicon for offensive tamazight language detection. In: 7th International Workshop on Social Media World Sensors. Sideways'22. Association for Computing Machinery, New York, NY, USA (2022). https://doi.org/10.1145/3544795.3544852. https://doi.org/10.1145/3544795.3544852

[54] Rizwan, H., Shakeel, M.H., Karim, A.: Hate-speech and offensive language detection in Roman Urdu. In: Proceedings of the 2020 Conference on Empirical Methods in Natural Language Processing (EMNLP), pp. 2512–2522. Association for Computational Linguistics, Online (2020). https://doi.org/10.18653/v1/2020.emnlp-main.197. https://aclanthology.org/2020.emnlp-main.197

[55] Sandaruwan, H.M.S.T., Lorensuhewa, S.A.S., Kalyani, M.A.L.: Sinhala hate speech detection in social media using text mining and machine learning. In: 2019 19th International Conference on Advances in ICT for Emerging Regions (ICTer), vol. 250, pp. 1–8 (2019). https://doi.org/10.1109/ICTer48817.2019.9023655

[56] Rathnayake, H., Sumanapala, J., Rukshani, R., Ranathunga, S.: Adapter-based fine-tuning of pre-trained multilingual language models for code-mixed and code-switched text classification. Knowledge and Information Systems **64**(7), 1937–1966 (2022). https://doi.org/10.1007/s10115-022-01698-1

[57] Naim, J., Hossain, T., Tasneem, F., Chy, A.N., Aono, M.: Leveraging fusion of sequence tagging models for toxic spans detection. Neurocomputing **500**, 688–702 (2022). https://doi.org/10.1016/j.neucom.2022.05.049

[58] Ribeiro, M.T., Singh, S., Guestrin, C.: "why should i trust you?": Explaining the predictions of any classifier. In: Proceedings of the 22nd ACM SIGKDD International Conference on Knowledge Discovery and Data Mining. KDD '16, pp. 1135–1144. Association for Computing Machinery, New York, NY, USA (2016). https://doi.org/10.1145/2939672.2939778. https://doi.org/10.1145/2939672.2939778

[59] Lei, T., Barzilay, R., Jaakkola, T.: Rationalizing neural predictions. In: Proceedings of the 2016 Conference on Empirical Methods in Natural Language Processing, pp. 107–117. Association for Computational Linguistics, Austin, Texas (2016). https://doi.org/10.18653/v1/D16-1011. https://aclanthology.org/D16-1011

[60] Da San Martino, G., Cresci, S., Barrón-Cedeño, A., Yu, S., Di Pietro, R., Nakov, P.: A survey on computational propaganda detection. In: Proceedings of the Twenty-Ninth International Joint Conference on Artificial Intelligence. IJCAI'20 (2021)

[61] Fomicheva, M., Specia, L., Aletras, N.: Translation error detection as rationale extraction. In: Findings of the Association for Computational Linguistics: ACL 2022, pp. 4148–4159. Association for Computational Linguistics, Dublin, Ireland (2022). https://doi.org/10.18653/v1/2022.findings-acl.327. https://aclanthology.org/2022.findings-acl.327

[62] Pavlopoulos, J., Sorensen, J., Laugier, L., Androutsopoulos, I.: SemEval-2021 task 5: Toxic spans detection. In: Proceedings of the 15th International Workshop on Semantic Evaluation (SemEval-2021), pp. 59–69. Association for Computational Linguistics, Online (2021). https://doi.org/10.18653/v1/2021.semeval-1.6. https://aclanthology.org/2021.semeval-1.6

[63] Dadvar, M., Trieschnigg, D., Ordelman, R., de Jong, F.: Improving cyberbullying detection with user context. In: Serdyukov, P., Braslavski, P., Kuznetsov, S.O., Kamps, J., Rüger, S., Agichtein, E., Segalovich, I., Yilmaz, E. (eds.) Advances in Information Retrieval, pp. 693–696. Springer, Berlin, Heidelberg (2013)

[64] Nobata, C., Tetreault, J., Thomas, A., Mehdad, Y., Chang, Y.: Abusive language detection in online user content. In: Proceedings of the 25th International Conference on World Wide Web. WWW '16, pp. 145–153. International World Wide Web Conferences Steering Committee, Republic and Canton of Geneva, CHE (2016). https://doi.org/10.1145/2872427.2883062. https://doi.org/10.1145/2872427.2883062

[65] Malmasi, S., Zampieri, M.: Detecting hate speech in social media. In: Proceedings of the International Conference Recent Advances in Natural Language Processing, 2017, pp. 467–472. INCOMA Ltd., Varna, Bulgaria (2017). https://doi.org/10.26615/978-954-452-049-6_062

[66] Mikolov, T., Sutskever, I., Chen, K., Corrado, G.S., Dean, J.: Distributed representations of words and phrases and their compositionality. In: Burges, C.J., Bottou, L., Welling, M., Ghahramani, Z., Weinberger, K.Q. (eds.) Advances in Neural Information Processing Systems, vol. 26. Curran Associates, Inc., ??? (2013)

[67] Pennington, J., Socher, R., Manning, C.: GloVe: Global vectors for word representation. In: Proceedings of the 2014 Conference on Empirical Methods in Natural Language Processing (EMNLP), pp. 1532–1543. Association for Computational Linguistics, Doha, Qatar (2014). https://doi.org/10.3115/v1/D14-1162. https://aclanthology.org/D14-1162

[68] Pavlopoulos, J., Malakasiotis, P., Androutsopoulos, I.: Deeper attention to abusive user content moderation. In: Proceedings of the 2017 Conference on Empirical Methods in Natural Language Processing, pp. 1125–1135. Association for Computational Linguistics, Copenhagen, Denmark (2017). https://doi.org/10.18653/v1/D17-1117. https://aclanthology.org/D17-1117

[69] Liu, G., Guo, J.: Bidirectional lstm with attention mechanism and convolutional layer for text classification. Neurocomputing **337**, 325–338 (2019). https://doi.org/10.1016/j.neucom.2019.01.078

[70] Aroyehun, S.T., Gelbukh, A.: Aggression detection in social media: Using deep neural networks, data augmentation, and pseudo labeling. In: Proceedings of the First Workshop on Trolling, Aggression and Cyberbullying (TRAC-2018), pp. 90–97. Association for Computational Linguistics, Santa Fe, New Mexico, USA (2018). https://aclanthology.org/W18-4411

[71] Peng, H., Li, J., He, Y., Liu, Y., Bao, M., Wang, L., Song, Y., Yang, Q.: Large-scale hierarchical text classification with recursively regularized deep graph-cnn. In: Proceedings of the 2018 World Wide Web Conference. WWW '18, pp. 1063–1072. International World Wide Web Conferences Steering Committee, Republic and Canton of Geneva, CHE (2018). https://doi.org/10.1145/3178876.3186005. https://doi.org/10.1145/3178876.3186005

[72] Pradhan, R., Chaturvedi, A., Tripathi, A., Sharma, D.K.: A review on offensive language detection. In: Kolhe, M.L., Tiwari, S., Trivedi, M.C., Mishra, K.K. (eds.) Advances in Data and Information Sciences, pp. 433–439. Springer, Singapore (2020)

[73] Hettiarachchi, H., Ranasinghe, T.: Emoji powered capsule network to detect type and target of offensive posts in social media. In: Proceedings of the International Conference on Recent Advances in Natural Language Processing (RANLP 2019), pp. 474–480. INCOMA Ltd., Varna, Bulgaria (2019). https://doi.org/10.26615/978-954-452-056-4_056. https://aclanthology.org/R19-1056

[74] Tang, X., Shen, X., Wang, Y., Yang, Y.: Categorizing offensive language in social networks: A chinese corpus, systems and an explanation tool. In: Sun, M., Li, S., Zhang, Y., Liu, Y., He, S., Rao, G. (eds.) Chinese Computational Linguistics, pp. 300–315. Springer, Cham (2020)

[75] Mishra, P., Del Tredici, M., Yannakoudakis, H., Shutova, E.: Abusive Language Detection with Graph Convolutional Networks. In: Proceedings of the 2019 Conference of the North American Chapter of the Association for Computational Linguistics: Human Language Technologies, Volume 1 (Long and Short Papers), pp. 2145–2150. Association for Computational Linguistics, Minneapolis, Minnesota (2019). https://doi.org/10.18653/v1/N19-1221. https://aclanthology.org/N19-1221

[76] Devlin, J., Chang, M.-W., Lee, K., Toutanova, K.: BERT: Pre-training of deep bidirectional transformers for language understanding. In: Proceedings of the 2019 Conference of the North American Chapter of the Association for Computational Linguistics: Human Language Technologies, Volume 1 (Long and Short Papers), pp. 4171–4186. Association for Computational Linguistics, Minneapolis, Minnesota (2019). https:

//doi.org/10.18653/v1/N19-1423. https://aclanthology.org/N19-1423

[77] Yang, Z., Dai, Z., Yang, Y., Carbonell, J., Salakhutdinov, R.R., Le, Q.V.: Xlnet: Generalized autoregressive pretraining for language understanding. In: Wallach, H., Larochelle, H., Beygelzimer, A., d'Alché-Buc, F., Fox, E., Garnett, R. (eds.) Advances in Neural Information Processing Systems, vol. 32. Curran Associates, Inc., ??? (2019). https://proceedings.neurips.cc/paper/2019/file/dc6a7e655d7e5840e66733e9ee67cc69-Paper.pdf

[78] Hettiarachchi, H., Ranasinghe, T.: BRUMS at SemEval-2020 task 3: Contextualised embeddings for predicting the (graded) effect of context in word similarity. In: Proceedings of the Fourteenth Workshop on Semantic Evaluation, pp. 142–149. International Committee for Computational Linguistics, Barcelona (online) (2020). https://doi.org/10.18653/v1/2020.semeval-1.16. https://aclanthology.org/2020.semeval-1.16

[79] Wiedemann, G., Yimam, S.M., Biemann, C.: UHH-LT at SemEval-2020 task 12: Fine-tuning of pre-trained transformer networks for offensive language detection. In: Proceedings of the Fourteenth Workshop on Semantic Evaluation, pp. 1638–1644. International Committee for Computational Linguistics, Barcelona (online) (2020). https://doi.org/10.18653/v1/2020.semeval-1.213. https://aclanthology.org/2020.semeval-1.213

[80] Risch, J., Krestel, R.: Bagging BERT models for robust aggression identification. In: Proceedings of the Second Workshop on Trolling, Aggression and Cyberbullying, pp. 55–61. European Language Resources Association (ELRA), Marseille, France (2020). https://aclanthology.org/2020.trac-1.9

[81] Sarkar, D., Zampieri, M., Ranasinghe, T., Ororbia, A.: fBERT: A neural transformer for identifying offensive content. In: Findings of the Association for Computational Linguistics: EMNLP 2021, pp. 1792–1798. Association for Computational Linguistics, Punta Cana, Dominican Republic (2021). https://doi.org/10.18653/v1/2021.findings-emnlp.154. https://aclanthology.org/2021.findings-emnlp.154

[82] Caselli, T., Basile, V., Mitrović, J., Granitzer, M.: HateBERT: Retraining BERT for abusive language detection in English. In: Proceedings of the 5th Workshop on Online Abuse and Harms (WOAH 2021), pp. 17–25. Association for Computational Linguistics, Online (2021). https://doi.org/10.18653/v1/2021.woah-1.3. https://aclanthology.org/2021.woah-1.3

[83] Conneau, A., Khandelwal, K., Goyal, N., Chaudhary, V., Wenzek, G., Guzmán, F., Grave, E., Ott, M., Zettlemoyer, L., Stoyanov,

V.: Unsupervised cross-lingual representation learning at scale. In: Proceedings of the 58th Annual Meeting of the Association for Computational Linguistics, pp. 8440–8451. Association for Computational Linguistics, Online (2020). https://doi.org/10.18653/v1/2020.acl-main.747. https://aclanthology.org/2020.acl-main.747

[84] Pfeiffer, J., Vulić, I., Gurevych, I., Ruder, S.: MAD-X: An Adapter-Based Framework for Multi-Task Cross-Lingual Transfer. In: Proceedings of the 2020 Conference on Empirical Methods in Natural Language Processing (EMNLP), pp. 7654–7673. Association for Computational Linguistics, Online (2020). https://doi.org/10.18653/v1/2020.emnlp-main.617. https://aclanthology.org/2020.emnlp-main.617

[85] Zhu, Q., Lin, Z., Zhang, Y., Sun, J., Li, X., Lin, Q., Dang, Y., Xu, R.: HITSZ-HLT at SemEval-2021 task 5: Ensemble sequence labeling and span boundary detection for toxic span detection. In: Proceedings of the 15th International Workshop on Semantic Evaluation (SemEval-2021), pp. 521–526. Association for Computational Linguistics, Online (2021). https://doi.org/10.18653/v1/2021.semeval-1.63. https://aclanthology.org/2021.semeval-1.63

[86] Ranasinghe, T., Sarkar, D., Zampieri, M., Ororbia, A.: WLV-RIT at SemEval-2021 task 5: A neural transformer framework for detecting toxic spans. In: Proceedings of the 15th International Workshop on Semantic Evaluation (SemEval-2021), pp. 833–840. Association for Computational Linguistics, Online (2021). https://doi.org/10.18653/v1/2021.semeval-1.111. https://aclanthology.org/2021.semeval-1.111

[87] Palomino, M., Grad, D., Bedwell, J.: GoldenWind at SemEval-2021 task 5: Orthrus - an ensemble approach to identify toxicity. In: Proceedings of the 15th International Workshop on Semantic Evaluation (SemEval-2021), pp. 860–864. Association for Computational Linguistics, Online (2021). https://doi.org/10.18653/v1/2021.semeval-1.115. https://aclanthology.org/2021.semeval-1.115

[88] Burtenshaw, B., Kestemont, M.: UAntwerp at SemEval-2021 task 5: Spans are spans, stacking a binary word level approach to toxic span detection. In: Proceedings of the 15th International Workshop on Semantic Evaluation (SemEval-2021), pp. 898–903. Association for Computational Linguistics, Online (2021). https://doi.org/10.18653/v1/2021.semeval-1.121. https://aclanthology.org/2021.semeval-1.121

[89] Rusert, J.: NLP_UIOWA at Semeval-2021 task 5: Transferring toxic sets to tag toxic spans. In: Proceedings of the 15th International Workshop on Semantic Evaluation (SemEval-2021), pp. 881–887. Association for Computational Linguistics, Online (2021). https://doi.org/10.18653/

v1/2021.semeval-1.119. https://aclanthology.org/2021.semeval-1.119

[90] Pluciński, K., Klimczak, H.: GHOST at SemEval-2021 task 5: Is explanation all you need? In: Proceedings of the 15th International Workshop on Semantic Evaluation (SemEval-2021), pp. 852–859. Association for Computational Linguistics, Online (2021). https://doi.org/10.18653/v1/2021.semeval-1.114. https://aclanthology.org/2021.semeval-1.114

[91] Xiang, T., MacAvaney, S., Yang, E., Goharian, N.: ToxCCIn: Toxic content classification with interpretability. In: Proceedings of the Eleventh Workshop on Computational Approaches to Subjectivity, Sentiment and Social Media Analysis, pp. 1–12. Association for Computational Linguistics, Online (2021). https://aclanthology.org/2021.wassa-1.1

[92] Karimi, A., Rossi, L., Prati, A.: UniParma at SemEval-2021 task 5: Toxic spans detection using CharacterBERT and bag-of-words model. In: Proceedings of the 15th International Workshop on Semantic Evaluation (SemEval-2021), pp. 220–224. Association for Computational Linguistics, Online (2021). https://doi.org/10.18653/v1/2021.semeval-1.25. https://aclanthology.org/2021.semeval-1.25

[93] Ranasinghe, T., Zampieri, M.: MUDES: Multilingual detection of offensive spans. In: Proceedings of the 2021 Conference of the North American Chapter of the Association for Computational Linguistics: Human Language Technologies: Demonstrations, pp. 144–152. Association for Computational Linguistics, Online (2021). https://doi.org/10.18653/v1/2021.naacl-demos.17. https://aclanthology.org/2021.naacl-demos.17

[94] Nouri, N.: Data augmentation with dual training for offensive span detection. In: Proceedings of the 2022 Conference of the North American Chapter of the Association for Computational Linguistics: Human Language Technologies, pp. 2569–2575. Association for Computational Linguistics, Seattle, United States (2022). https://doi.org/10.18653/v1/2022.naacl-main.185. https://aclanthology.org/2022.naacl-main.185

[95] Sutton, C., McCallum, A.: An introduction to conditional random fields. Found. Trends Mach. Learn. **4**(4), 267–373 (2012). https://doi.org/10.1561/2200000013

[96] Yan, E., Tayyar Madabushi, H.: UoB at SemEval-2021 task 5: Extending pre-trained language models to include task and domain-specific information for toxic span prediction. In: Proceedings of the 15th International Workshop on Semantic Evaluation (SemEval-2021), pp. 243–248. Association for Computational Linguistics, Online (2021). https://doi.org/10.18653/v1/2021.semeval-1.28. https://aclanthology.org/2021.semeval-1.28

[97] Paraschiv, A., Cercel, D.-C., Dascalu, M.: UPB at SemEval-2021 task 5: Virtual adversarial training for toxic spans detection. In: Proceedings of the 15th International Workshop on Semantic Evaluation (SemEval-2021), pp. 225–232. Association for Computational Linguistics, Online (2021). https://doi.org/10.18653/v1/2021.semeval-1.26. https://aclanthology.org/2021.semeval-1.26

[98] DeYoung, J., Jain, S., Rajani, N.F., Lehman, E., Xiong, C., Socher, R., Wallace, B.C.: ERASER: A benchmark to evaluate rationalized NLP models. In: Proceedings of the 58th Annual Meeting of the Association for Computational Linguistics, pp. 4443–4458. Association for Computational Linguistics, Online (2020). https://doi.org/10.18653/v1/2020.acl-main.408. https://aclanthology.org/2020.acl-main.408

[99] Jain, S., Wiegreffe, S., Pinter, Y., Wallace, B.C.: Learning to faithfully rationalize by construction. In: Proceedings of the 58th Annual Meeting of the Association for Computational Linguistics, pp. 4459–4473. Association for Computational Linguistics, Online (2020). https://doi.org/10.18653/v1/2020.acl-main.409. https://aclanthology.org/2020.acl-main.409

[100] Lundberg, S.M., Lee, S.-I.: A unified approach to interpreting model predictions. NIPS'17, pp. 4768–4777. Curran Associates Inc., Red Hook, NY, USA (2017)

[101] Taleb, M., Hamza, A., Zouitni, M., Burmani, N., Lafkiar, S., En-Nahnahi, N.: Detection of toxicity in social media based on natural language processing methods. In: 2022 International Conference on Intelligent Systems and Computer Vision (ISCV), pp. 1–7 (2022). https://doi.org/10.1109/ISCV54655.2022.9806096

[102] Ding, H., Jurgens, D.: HamiltonDinggg at SemEval-2021 task 5: Investigating toxic span detection using RoBERTa pre-training. In: Proceedings of the 15th International Workshop on Semantic Evaluation (SemEval-2021), pp. 263–269. Association for Computational Linguistics, Online (2021). https://doi.org/10.18653/v1/2021.semeval-1.31. https://aclanthology.org/2021.semeval-1.31

[103] Waseem, Z., Hovy, D.: Hateful symbols or hateful people? predictive features for hate speech detection on Twitter. In: Proceedings of the NAACL Student Research Workshop, pp. 88–93. Association for Computational Linguistics, San Diego, California (2016). https://doi.org/10.18653/v1/N16-2013. https://aclanthology.org/N16-2013

[104] Zampieri, M., Malmasi, S., Nakov, P., Rosenthal, S., Farra, N., Kumar, R.: Predicting the type and target of offensive posts in social media. In: Proceedings of the 2019 Conference of the North American Chapter of

the Association for Computational Linguistics: Human Language Technologies, Volume 1 (Long and Short Papers), pp. 1415–1420. Association for Computational Linguistics, Minneapolis, Minnesota (2019). https://doi.org/10.18653/v1/N19-1144. https://aclanthology.org/N19-1144

[105] Burnap, P., Williams, M.L.: Cyber hate speech on twitter: An application of machine classification and statistical modeling for policy and decision making. Policy & Internet **7**(2), 223–242 (2015). https://doi.org/10.1002/poi3.85

[106] Perry, T.: LightTag: Text annotation platform. In: Proceedings of the 2021 Conference on Empirical Methods in Natural Language Processing: System Demonstrations, pp. 20–27. Association for Computational Linguistics, Online and Punta Cana, Dominican Republic (2021). https://doi.org/10.18653/v1/2021.emnlp-demo.3. https://aclanthology.org/2021.emnlp-demo.3

[107] Lhoest, Q., Villanova del Moral, A., Jernite, Y., Thakur, A., von Platen, P., Patil, S., Chaumond, J., Drame, M., Plu, J., Tunstall, L., Davison, J., Šaško, M., Chhablani, G., Malik, B., Brandeis, S., Le Scao, T., Sanh, V., Xu, C., Patry, N., McMillan-Major, A., Schmid, P., Gugger, S., Delangue, C., Matussière, T., Debut, L., Bekman, S., Cistac, P., Goehringer, T., Mustar, V., Lagunas, F., Rush, A., Wolf, T.: Datasets: A community library for natural language processing. In: Proceedings of the 2021 Conference on Empirical Methods in Natural Language Processing: System Demonstrations, pp. 175–184. Association for Computational Linguistics, Online and Punta Cana, Dominican Republic (2021). https://doi.org/10.18653/v1/2021.emnlp-demo.21. https://aclanthology.org/2021.emnlp-demo.21

[108] Schwarm, S., Ostendorf, M.: Reading level assessment using support vector machines and statistical language models. In: Proceedings of the 43rd Annual Meeting of the Association for Computational Linguistics (ACL'05), pp. 523–530. Association for Computational Linguistics, Ann Arbor, Michigan (2005). https://doi.org/10.3115/1219840.1219905. https://aclanthology.org/P05-1065

[109] Goudjil, M., Koudil, M., Bedda, M., Ghoggali, N.: A novel active learning method using svm for text classification. International Journal of Automation and Computing **15**(3), 290–298 (2018). https://doi.org/10.1007/s11633-015-0912-z

[110] Alakrot, A., Murray, L., Nikolov, N.S.: Towards accurate detection of offensive language in online communication in arabic. Procedia Computer Science **142**, 315–320 (2018). https://doi.org/10.1016/j.procs.2018.10.491. Arabic Computational Linguistics

[111] Kim, Y.: Convolutional neural networks for sentence classification. In: Proceedings of the 2014 Conference on Empirical Methods in Natural Language Processing (EMNLP), pp. 1746–1751. Association for Computational Linguistics, Doha, Qatar (2014). https://doi.org/10.3115/v1/D14-1181. https://aclanthology.org/D14-1181

[112] Bojanowski, P., Grave, E., Joulin, A., Mikolov, T.: Enriching Word Vectors with Subword Information. Transactions of the Association for Computational Linguistics **5**, 135–146 (2017) https://direct.mit.edu/tacl/article-pdf/doi/10.1162/tacl_a_00051/1567442/tacl_a_00051.pdf. https://doi.org/10.1162/tacl_a_00051

[113] Lakmal, D., Ranathunga, S., Peramuna, S., Herath, I.: Word embedding evaluation for Sinhala. In: Proceedings of the 12th Language Resources and Evaluation Conference, pp. 1874–1881. European Language Resources Association, Marseille, France (2020). https://aclanthology.org/2020.lrec-1.231

[114] Wolf, T., Debut, L., Sanh, V., Chaumond, J., Delangue, C., Moi, A., Cistac, P., Rault, T., Louf, R., Funtowicz, M., Davison, J., Shleifer, S., von Platen, P., Ma, C., Jernite, Y., Plu, J., Xu, C., Le Scao, T., Gugger, S., Drame, M., Lhoest, Q., Rush, A.: Transformers: State-of-the-art natural language processing. In: Proceedings of the 2020 Conference on Empirical Methods in Natural Language Processing: System Demonstrations, pp. 38–45. Association for Computational Linguistics, Online (2020). https://doi.org/10.18653/v1/2020.emnlp-demos.6. https://aclanthology.org/2020.emnlp-demos.6

[115] Dhananjaya, V., Demotte, P., Ranathunga, S., Jayasena, S.: BERTifying Sinhala - a comprehensive analysis of pre-trained language models for Sinhala text classification. In: Proceedings of the Thirteenth Language Resources and Evaluation Conference, pp. 7377–7385. European Language Resources Association, Marseille, France (2022). https://aclanthology.org/2022.lrec-1.803

[116] Barbieri, F., Espinosa Anke, L., Camacho-Collados, J.: XLM-T: Multilingual language models in Twitter for sentiment analysis and beyond. In: Proceedings of the Thirteenth Language Resources and Evaluation Conference, pp. 258–266. European Language Resources Association, Marseille, France (2022). https://aclanthology.org/2022.lrec-1.27

[117] Bansal, T., Jha, R., McCallum, A.: Learning to few-shot learn across diverse natural language classification tasks. In: Proceedings of the 28th International Conference on Computational Linguistics, pp. 5108–5123. International Committee on Computational Linguistics, Barcelona,

Spain (Online) (2020). https://doi.org/10.18653/v1/2020.coling-main.448. https://aclanthology.org/2020.coling-main.448

[118] Lafferty, J.D., McCallum, A., Pereira, F.C.N.: Conditional random fields: Probabilistic models for segmenting and labeling sequence data. In: Proceedings of the Eighteenth International Conference on Machine Learning. ICML '01, pp. 282–289. Morgan Kaufmann Publishers Inc., San Francisco, CA, USA (2001)

[119] K, K., Wang, Z., Mayhew, S., Roth, D.: Cross-lingual ability of multilingual bert: An empirical study. In: International Conference on Learning Representations (2020). https://openreview.net/forum?id=HJeT3yrtDr

[120] Gair, J.W.: Sinhala, an indo-aryan isolate. South Asian Review **6**(3), 51–64 (1982) https://doi.org/10.1080/02759527.1982.11933091. https://doi.org/10.1080/02759527.1982.11933091

[121] Zhou, Y., Goldman, S.: Democratic co-learning. In: 16th IEEE International Conference on Tools with Artificial Intelligence, pp. 594–602 (2004). https://doi.org/10.1109/ICTAI.2004.48

[122] Hettiarachchi, H., Al-Turkey, D., Adedoyin-Olowe, M., Bhogal, J., Gaber, M.M.: Ted-s: Twitter event data in sports and politics with aggregated sentiments. Data **7**(7) (2022). https://doi.org/10.3390/data7070090

[123] Mohamed, T.A., El Gayar, N., Atiya, A.F.: A co-training approach for time series prediction with missing data. In: Haindl, M., Kittler, J., Roli, F. (eds.) Multiple Classifier Systems, pp. 93–102. Springer, Berlin, Heidelberg (2007)

[124] Gou, J., Yu, B., Maybank, S.J., Tao, D.: Knowledge distillation: A survey. International Journal of Computer Vision **129**(6), 1789–1819 (2021). https://doi.org/10.1007/s11263-021-01453-z

[125] Yu, S., Kulkarni, N., Lee, H., Kim, J.: On-device neural language model based word prediction. In: Proceedings of the 27th International Conference on Computational Linguistics: System Demonstrations, pp. 128–131. Association for Computational Linguistics, Santa Fe, New Mexico (2018). https://aclanthology.org/C18-2028

[126] Guo, D., Kim, Y., Rush, A.: Sequence-level mixed sample data augmentation. In: Proceedings of the 2020 Conference on Empirical Methods in Natural Language Processing (EMNLP), pp. 5547–5552. Association for Computational Linguistics, Online (2020). https://doi.org/10.18653/v1/2020.emnlp-main.447. https://aclanthology.org/2020.emnlp-main.447

[127] Gajbhiye, A., Fomicheva, M., Alva-Manchego, F., Blain, F., Obamuyide, A., Aletras, N., Specia, L.: Knowledge distillation for quality estimation. In: Findings of the Association for Computational Linguistics: ACL-IJCNLP 2021, pp. 5091–5099. Association for Computational Linguistics, Online (2021). https://doi.org/10.18653/v1/2021.findings-acl.452. https://aclanthology.org/2021.findings-acl.452

[128] Zhou, X., Zhang, X., Tao, C., Chen, J., Xu, B., Wang, W., Xiao, J.: Multi-grained knowledge distillation for named entity recognition. In: Proceedings of the 2021 Conference of the North American Chapter of the Association for Computational Linguistics: Human Language Technologies, pp. 5704–5716. Association for Computational Linguistics, Online (2021). https://doi.org/10.18653/v1/2021.naacl-main.454. https://aclanthology.org/2021.naacl-main.454